\documentclass[10pt,twocolumn,letterpaper]{article}

\usepackage[pagenumbers]{cvpr} % To force page numbers, e.g. for an arXiv version

\usepackage{graphicx}
\usepackage{amsmath}
\usepackage{amssymb}
\usepackage{booktabs}
\usepackage{epsfig}
\usepackage{graphicx}
\usepackage{float}
\usepackage{caption}
\usepackage{pifont}
\usepackage{array}
\usepackage{arydshln}
\usepackage{multirow, makecell}
\usepackage{siunitx}
\usepackage{lipsum}
\usepackage{rotating}
\usepackage{csquotes}
\usepackage{algorithm}
\usepackage{algpseudocode}
\usepackage[dvipsnames]{xcolor}
\usepackage{etoolbox}

    \makeatletter
\def\@fnsymbol#1{\ensuremath{\ifcase#1\or \dagger\or \ddagger\or
   \mathsection\or \mathparagraph\or \|\or **\or \dagger\dagger
   \or \ddagger\ddagger \else\@ctrerr\fi}}
    \makeatother

\usepackage{subcaption}
\usepackage[pagebackref,breaklinks,colorlinks]{hyperref}

\newcommand{\parheader}{\smallskip\noindent\textbf}
\newcommand{\losslambda}{privacy parameter\xspace}%this is subject to change
\newcommand{\privacy}{content concealment\xspace}%this is subject to change
\newcommand{\Privacy}{Content concealment\xspace}%this is subject to change
\newcommand{\privacying}{content-concealing\xspace}%this is subject to change
\usepackage[capitalize]{cleveref}
\crefname{section}{Sec.}{Secs.}
\Crefname{section}{Section}{Sections}
\Crefname{table}{Table}{Tables}
\crefname{table}{Tab.}{Tabs.}
\Crefname{fig}{Figure}{Figures}
\crefname{fig}{Fig.}{Figs.}

\begin{document}

%%%%%%%%% TITLE - PLEASE UPDATE
\title{NinjaDesc: Content-Concealing Visual Descriptors via Adversarial Learning}

\author{Tony Ng\textsuperscript{1,2}
% For a paper whose authors are all at the same institution,
% omit the following lines up until the closing ``}''.
% Additional authors and addresses can be added with ``\and'',
% just like the second author.
% To save space, use either the email address or home page, not both
\hspace{6pt}
%\and
Hyo Jin Kim\textsuperscript{1}\thanks{Corresponding author.}
\hspace{6pt}
%\and
Vincent T. Lee\textsuperscript{1}
\hspace{6pt}
%\and
Daniel DeTone\textsuperscript{1}
\hspace{6pt}
%\and
Tsun-Yi Yang\textsuperscript{1}
%\hspace{6pt}
%\and
\\
Tianwei Shen\textsuperscript{1}\hspace{6pt}
%\and
Eddy Ilg\textsuperscript{1}\hspace{6pt}
%\and
Vassileios Balntas\textsuperscript{1}\hspace{6pt}
%\and
Krystian Mikolajczyk\textsuperscript{2}\hspace{6pt}
%\and
Chris Sweeney\textsuperscript{1}
%\and
\vspace{.6ex}
\\
\textsuperscript{1}Reality Labs, Meta
%\and
\hspace{12pt}
\textsuperscript{2}Imperial College London
}
\maketitle

\vspace{-0pt}
%%%%%%%%% ABSTRACT
\begin{abstract}
    In the light of recent analyses on privacy-concerning scene revelation from visual descriptors, we develop descriptors that conceal the input image content. In particular, we propose an adversarial learning framework for training visual descriptors that prevent image reconstruction, while maintaining the matching accuracy. We let a feature encoding network and image reconstruction network compete with each other, such that the feature encoder tries to impede the image reconstruction with its generated descriptors, while the reconstructor tries to recover the input image from the descriptors. The experimental results demonstrate that the visual descriptors obtained with our method significantly deteriorate the image reconstruction quality with minimal impact on correspondence matching and camera localization performance. 
    %matchability
\end{abstract}

\vspace{-5pt}
%%%%%%%%% ABSTRACT
%%%%%%%%% BODY TEXT
\section{Introduction}
\label{sec:intro}

% Visual descriptors are used everywhere
Local visual descriptors \cite{detone2018superpoint,tian2020hynet,barroso-laguna2019keynet,revaud2019r2d2,tian2019sosnet} are fundamental to a wide range of computer vision applications such as SLAM \cite{mur-artal2016orb-slam,newcombe2011dtam,mei2011rslam,dong2015distributed}, SfM \cite{schonberger2016colmap,sweeney2015theia,agarwal2011building}, wide-baseline stereo~\cite{jin2021image, mur2017orb}, calibration~\cite{Oth_2013_CVPR}, tracking~\cite{hare2012efficient,nebehay2014consensus,pernici2013object}, image retrieval~\cite{tolias2020learning,ng2020solar,simeoni2019local,arandjelovic2016netvlad,kim2017learned,noh2017delf,arandjelovic2014dislocation,tolias2013smk}, and camera pose estimation~\cite{toft2020long,sattler2017active,porav2018adversarial,sarlin20superglue,dusmanu2019d2net,baik2020domain,toft2020single}. These descriptors represent local regions of images and are used to establish local correspondences between and across images and 3D models. 

% Recent analysis on revealing scenes
The descriptors take the form of vectors in high-dimensional space, and thus are not directly interpretable by humans. However, researchers have shown that it is possible to reveal the input images from local visual descriptors~\cite{weinzaepfel2011reconstructing,dosovitskiy2016inverting, d2013bits}. With the recent advances in deep learning, the quality of the reconstructed image content has been significantly improved~\cite{invsfm,dangwal21}. This poses potential privacy concerns for visual descriptors if they are used for sensitive data without  proper encryption \cite{dangwal21,linecloud,weinzaepfel2011reconstructing}. % Added for the legal review %\ty{Image reconstruction/completion/inpainting and descriptor visual inversion are several different tasks. Only the latter one raises security concern, and many method do not rely on adversarial process.} %/hk{I meant adversarial learning as in InvSfM}

% How others try to address this and their limitations
To prevent the reconstruction of the image content from visual descriptors, several methods have been proposed. These methods include obfuscating keypoint locations by lifting them to lines that pass through the original points~\cite{speciale2019privacy,linecloud,geppert2021privacy,shibuya2020privacy}, or to affine subspaces with augmented adversarial feature samples~\cite{dusmanu2020privacy} to increase the difficulty of recovering the original images. However, recent work \cite{chelani2021howprivacypreserving} has demonstrated that the closest points between lines can yield a good approximation to the original points' locations.
%, allowing descriptor inversion.

\begin{figure}[!t]
    \vspace{-0pt}
    \centering
    \includegraphics[width=\linewidth, trim={
        0pt, 0pt, 0pt, 0pt}, clip]{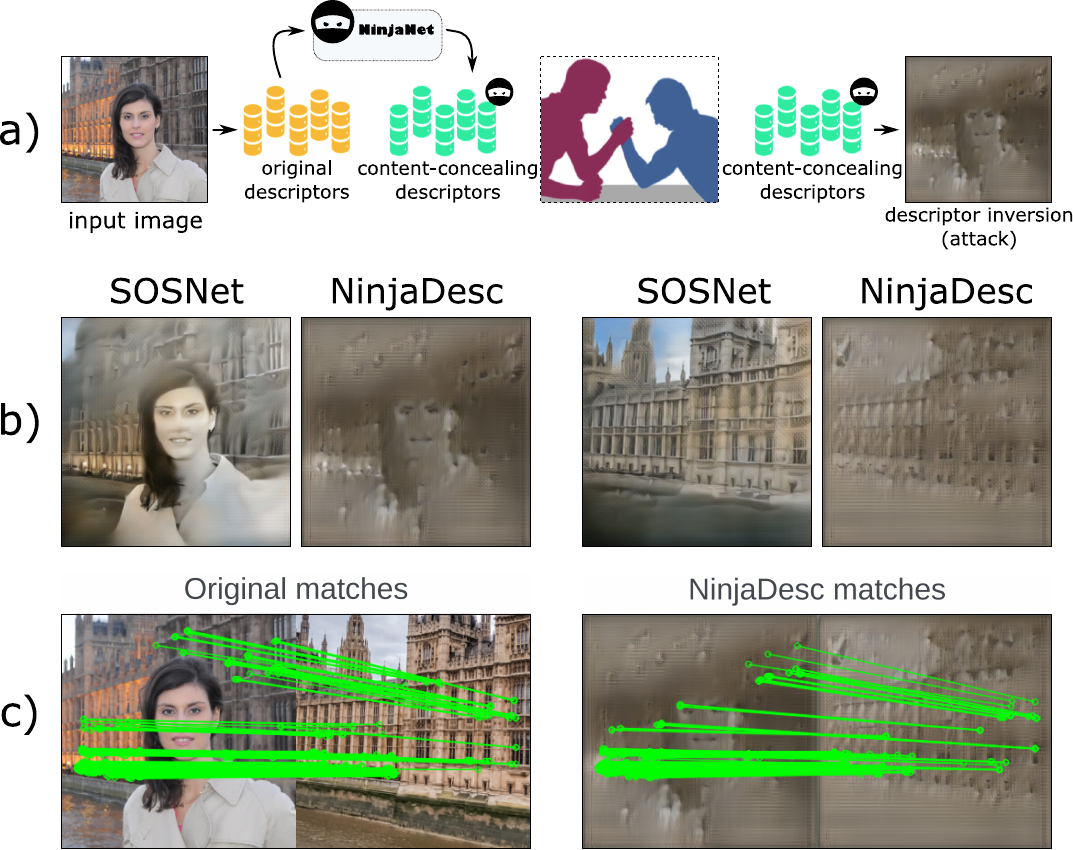}
    \vspace{-18pt}
    \caption{Our proposed \privacying visual descriptor. \textbf{a)} We train \textbf{\textit{NinjaNet}}, the \privacying network via adversarial learning to give \textbf{\textit{NinjaDesc}}. \textbf{b)} On the two examples shown, we compare inversions on SOSNet~\cite{tian2019sosnet} descriptors \vs NinjaDesc (encoding SOSNet with NinjaNet). \textbf{c)} NinjaDesc is able to conceal facial features and landmark structures, while retaining correspondences. 
    \textit{Image credits: \href{https://www.flickr.com/photos/laylamoran4battersea/4543311974/}{\textit{laylamoran4battersea}} \& \href{https://www.flickr.com/photos/sgerner/24010777130/}{\textit{sgerner}} (Flickr)}%
    \protect\footnotemark
    .
    }
    \label{fig:teaser}
    \vspace{-12pt}
\end{figure}
\footnotetext{CC BY 2.0 \& CC BY-SA 2.0 licenses.}

% How we approach the problem
In this work, we explore whether such local feature inversion could be mitigated at the descriptor level. Ideally, we want a descriptor that does not reveal the image content without a compromise in its performance.
This may seem counter-intuitive due to the trade-off between utility and privacy discussed in the recent analysis on visual descriptors \cite{dangwal21}, where the utility is defined as matching accuracy, and the privacy is defined as non-invertibility of the descriptors.
The analysis showed that the more useful the descriptors are for correspondence matching, the easier it is to invert them. 
To minimize this trade-off, we propose an adversarial approach to train visual descriptors. 

Specifically, we optimize our descriptor encoding network with an adversarial loss for descriptor invertibility, in addition to the traditional metric learning loss for feature correspondence matching. For the adversarial loss, we jointly train an image reconstruction network to compete with the descriptor network in revealing the original image content from the descriptors. In this way, the descriptor network learns to hinder the reconstruction network by generating visual descriptors that conceal the image content, while being optimized for correspondence matching. 

In particular, we introduce an auxiliary encoder network \textit{\textbf{NinjaNet}} that can be trained with any existing visual descriptors and transform them to our content-concealing \textit{\textbf{NinjaDesc}}, as illustrated in Fig.~\ref{fig:teaser}. In the experiments, we show that visual descriptors trained with our adversarial learning framework lead to only marginal drop in performance for feature matching and visual localization tasks, while significantly reducing the visual similarity of the reconstruction to the original input image.  

%In this work, we explore whether such local feature inversion could be mitigated at the descriptor level. In particular, we propose to take an adversarial approach to train visual descriptors to minimize the trade-off between utility and non-invertibility. 

% Benefits of our approach
One of the main benefits of our method is that we can control the trade-off between utility and privacy by changing a single parameter in the loss function.
In addition, our method generalizes to different types of visual descriptors, and different image reconstruction network architectures.

% Summary of contributions
In summary, our main innovations are as follows:
\noindent{\bf a)}~We propose a novel adversarial learning framework for visual descriptors to prevent reconstructing original input image content from the descriptors. We experimentally validate that the obtained descriptors significantly deteriorate the image quality from descriptor inversion with only marginal drop in matching accuracy using standard benchmarks for matching (HPatches~\cite{balntas2017hpatches}) and visual localization (Aachen Day-Night \cite{sattler2018benchmarking, zhang2020aachenv_1_1}). 
{\bf b)} We empirically demonstrate that we can effectively control the trade-off between utility (matching accuracy) and privacy (non-invertibility)  by changing a single training parameter. %that weighs the adversarial loss.  
\noindent{\bf c)} We provide ablation studies by using different types of visual descriptors, image reconstruction network architectures and scene categories to demonstrate the generalizability of our method. \\
\vspace{-5pt}
\section{Related work}
\label{sec:related-work}
%In this section, we first briefly discuss the past literature on visual descriptors and their applications on visual localization. Then, we go through the recent advances in privacy of visual localization methods, including both those that attack privacy and that defends against such attacks.

% \subsection{Local feature descriptors}
% \label{subsec:related_local-descs}

% \subsection{Privacy in visual localization}
% \label{subsec:related_privacy-in-vis-loc}
% \parheader{Privacy attack.}

% \parheader{Privacy preserving methods.}
% \lipsum[4-6]

This section discusses prior work on visual descriptor inversion and the state-of-the-art descriptor designs that attempt to prevent such inversion. 

\parheader{Inversion of visual descriptors.}
Early results of reconstructing images from local descriptors was shown by Weinzaepfel \etal~\cite{weinzaepfel2011reconstructing} by stitching the image patches from a known database with the closest distance to the input SIFT~\cite{lowe2004sift} descriptors in the feature space. d'Angelo \etal~\cite{d2013bits} used a deconvolution approach on local binary descriptors such as BRIEF~\cite{calonder2010brief} and FREAK~\cite{alahi2012freak}.
Vondrick \etal~\cite{hoggles} used paired dictionary learning to invert HoG~\cite{zhu2006fast} features to reveal its limitations for object detection. 
For global descriptors, Kato and Harada~\cite{kato2014image} reconstructed images from Bag-of-Words descriptors~\cite{sivic2003video}. % by optimizing the arrangement of visual words.
However, the quality of reconstructions by these early works were not sufficient to raise concerns about privacy or security.

Subsequent work introduced methods that steadily improved the quality of the reconstructions.
Mahendran and Vedaldi~\cite{mahendran2015understanding} used a back-propagation technique with a natural image prior to invert CNN features as well as SIFT~\cite{liu2010sift} and HOG~\cite{zhu2006fast}. Dosovitskiy and Brox~\cite{dosovitskiy2016inverting} trained up-convolutional networks that estimate the input image from features in a regression fashion, and demonstrated superior results on both classical~\cite{lowe2004sift,zhu2006fast,ojala2002multiresolution} and CNN~\cite{krizhevsky2017imagenet} features.
In the recent work, descriptor inversion methods have started to leverage larger and more advanced CNN models as well as employ advanced optimization techniques.
Pittaluga \etal~\cite{invsfm} and Dangwal \etal~\cite{dangwal21} demonstrated sufficiently high reconstruction qualities, revealing not only semantic information but also details in the original images. 
%demonstrated surprisingly good reconstruction quality from reverse engineering attacks using.
%Both works demonstrated sufficiently high reconstruction quality which can reveal not only semantic information but also details in the original input imagery. 

\parheader{Preventing descriptor inversion for privacy.} 
% Privacy implication
Descriptor inversion raises privacy concerns \cite{invsfm,dangwal21,linecloud,weinzaepfel2011reconstructing}.
For example, in computer vision systems where the visual descriptors are transferred between the device and the server, an honest-but-curious server may exploit the descriptors sent by the client device. 
In particular, many large-scale localization systems adopt cloud computing and storage, due to limited compute on mobile devices.
Homomorphic encryption \cite{erkin2009privacy,sadeghi2009efficient,yonetani2017privacy} can protect descriptors, but are too computationally expensive for large-scale applications.

% Mitigation methods: Line cloud, etc. etc.
Proposed by Speciale~\etal\cite{linecloud}, the line-cloud representation obfuscate 2D\,/\,3D point locations in the map building process~\cite{geppert2020privacy,shibuya2020privacy,geppert2021privacy} without compromising the accuracy in localization.
However, since the descriptors are unchanged, Chelani \etal~\cite{chelani2021howprivacypreserving} showed that line-clouds are vulnerable to inversion attacks if the underlying point-cloud is recovered.

Adversarial learning has been applied in image encoding~\cite{pittaluga2019learning,hinojosa2021learning,xiao2020adversarial} that optimizes privacy-utility trade-off, but not in the context of descriptor inversion, which involves reconstruction of images from a set of local features and has a much broader scope of downstream applications.

% TODO: Feature lifting paper that came out this year at CVPR
Recently, Dusmanu~\etal~\cite{dusmanu2020privacy} proposed a privacy-preserving visual descriptor via lifting descriptors to affine subspaces, which conceals the visual content from inversion attacks. However, this comes with a significant cost on the descriptor's utility in downstream tasks.
Our work differs from \cite{dusmanu2020privacy} in that we propose a learned \privacying descriptor and explicitly train it for utility retention to achieve a better trade-off between the two.
% TODO: This work is different from the above because of X, Y, and Z on top of this

% TODO: Not entirely sure the breakdown/taxonomization of the prior work here
%Other prior works try to alleviate the descriptor inversion concern by perturbing the images~\cite{ren2018learning, Li_2019_CVPR_Workshops,butler2015privacy,ryoo2016privacy,raval2017protecting,wu2018towards,pittaluga2019learning,Wang_2019_CVPR_Workshops}.
%Prior work~\cite{vishwamitra2017blur,ren2018learning, Li_2019_CVPR_Workshops, dangwal21} proposes either selective data suppression or substitution to blind parts of the image which may contain private information.
%Other approaches propose different encoding schemes or reducing image quality to minimize private image content~\cite{butler2015privacy,ryoo2016privacy,raval2017protecting,wu2018towards,pittaluga2019learning,Wang_2019_CVPR_Workshops}.
%Finally, an emerging class of techniques proposes cryptographic methods like homomorphic encryption \cite{erkin2009privacy,sadeghi2009efficient,yonetani2017privacy} to protect descriptors but are currently too computationally expensive for localization use cases.
%These approaches are complementary to the work we propose around engineering content-concealing descriptor and many of these prior approaches can be layered on top of ours. % TODO: validate this sentence

% TODO:
% \input{sections/3threat_model}
\section{Method}
\label{sec:method}
\begin{figure}[!t]
    \vspace{-0pt}
    \centering 
    \includegraphics[width=\linewidth, trim={
        0pt, 0pt, 0pt, 0pt}]{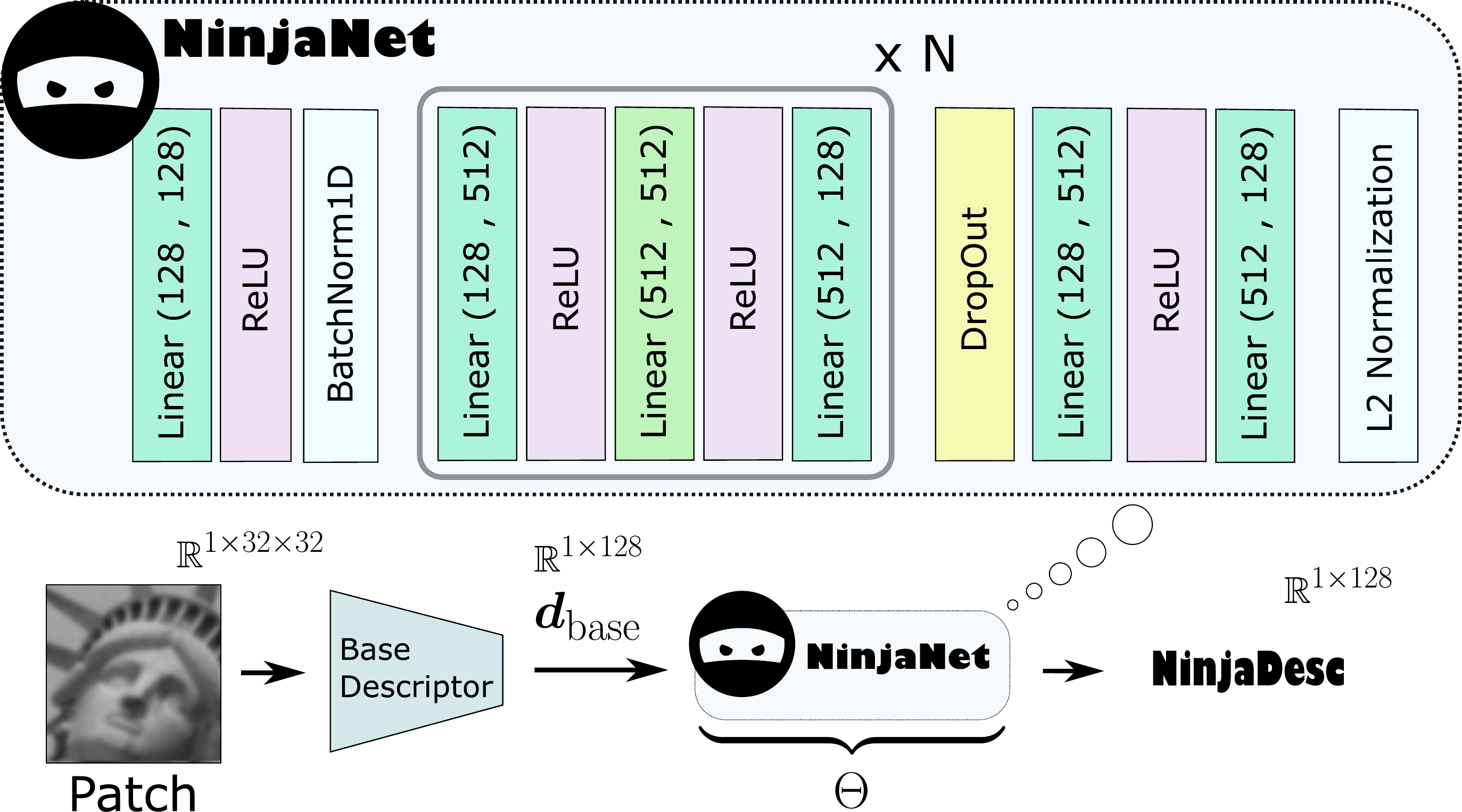}
    \vspace{-15pt}
    \caption{\textbf{Top:} Architecture of our \privacying \textbf{\textit{NinjaNet} encoder $\Theta$}. \textbf{Bottom:} A base descriptor with dimensionality $C$ is transformed to  \textbf{\textit{NinjaDesc}} of the same size, \eg $C=128$.}
    \label{fig:encoder}
    \vspace{-10pt}
\end{figure}

We propose an adversarial learning framework for obtaining content-concealing visual descriptors, by introducing a descriptor inversion model as an adversary. 
In this section, we detail our content-concealing encoder NinjaNet (Sec.~\ref{subsec:privacy_encoder}) and the descriptor inversion model (Sec.~\ref{subsec:inversion_model}), as well as the joint adversarial training procedure (Sec.~\ref{subsec:joint_adversarial_training}). 
\subsection{NinjaNet: the content-concealing encoder}
\label{subsec:privacy_encoder}
In order to conceal the visual content of a local descriptor while maintaining its utility, we need a trainable encoder which transforms the original descriptor space to a different one, where visual information essential for reconstruction is reduced.
Our  NinjaNet encoder $\Theta$ is implemented by an MLP shown in Fig. \ref{fig:encoder}. 
It takes a base descriptor $\boldsymbol{d}_{\text{base}}$, and transforms it into a content-concealing NinjaDesc, $\boldsymbol{d}_{\text{ninja}}$:
\begin{equation}
\boldsymbol{d}_{\text{ninja}} = \Theta(\boldsymbol{d}_{\text{base}})
\end{equation}
The design of NinjaNet is light-weight and plug-and-play, to make it flexible in accepting different types of existing local descriptors. The encoded NinjaDesc descriptor maintains the matching performance of the original descriptor, but prevents from high-quality reconstruction of images.
In many of our experiments, we adopt SOSNet~\cite{tian2019sosnet} as our base descriptor since it is one of the top-performing descriptors for correspondence matching and visual localization~\cite{jin2021image}.

\parheader{Utility initialization.}
To maintain the utility (\ie accuracy for downstream tasks) of our encoded descriptor, we use a patch-based descriptor training approach~\cite{tian2017l2net,mishchuk2017hardnet,tian2019sosnet}.
The initialization step trains NinjaNet via a triplet-based ranking loss.
We use the UBC dataset~\cite{goesele2007ubc} which contains three subsets of patches labelled as positive and negative pairs, allowing for easy implementation of triplet-loss training.

%%%%%%%%%%%%%%%%%%%%%%%%%%%%%%%%%%%%%%%%%%%%%%%%%%%%%%%%%%%%%%%
%%%%%%%%%%%%%%%%%%%%%%%%%%%%%%%%%%%%%%%%%%%%%%%%%%%%%%%%%%%%%%%
\begin{figure*}[!ht]
    \vspace{-0pt}
    \centering
    \includegraphics[width=\linewidth, trim={
        0pt, 0pt, 0pt, 0pt}]{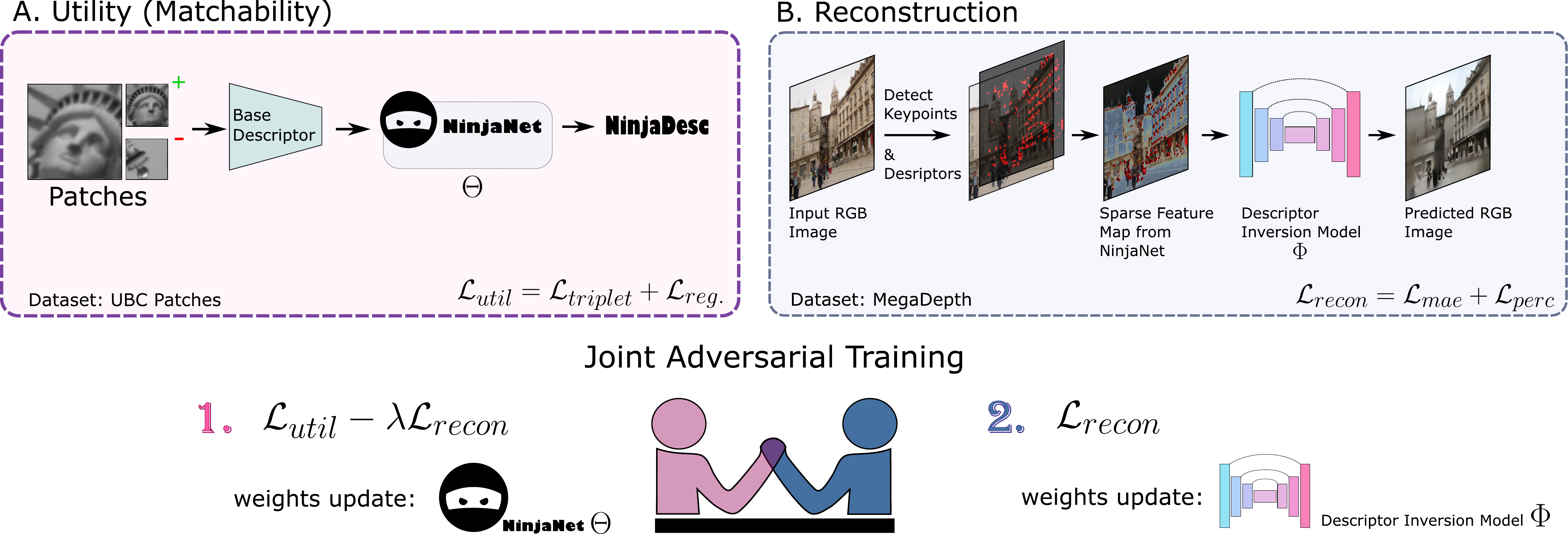}
    \vspace{-15pt}
    \caption{The pipeline for training our content-concealing NinjaDesc. 
        \textbf{Top:} The two networks at play and their corresponding objectives are: \textcolor{OrangeRed}{\textbf{1.}} NinjaNet $\Theta$, which is for utility retention in A; and \textcolor{NavyBlue}{\textbf{2.}} the descriptor inversion model, which reconstructs RGB images from input sparse features in B. 
        \textbf{Bottom:} During joint adversarial training, we alternate between steps \textcolor{OrangeRed}{\textbf{1.}} and \textcolor{NavyBlue}{\textbf{2.}}, which is presented by Algorithm~\ref{alg:joint_training}.
    }
    \label{fig:joint_training}
    \vspace{-5pt}
\end{figure*}

%%%%%%%%%%%%%%%%%%%%%%%%%%%%%%%%%%%%%%%%%%%%%%%%%%%%%%%%%%%%%%%
%%%%%%%%%%%%%%%%%%%%%%%%%%%%%%%%%%%%%%%%%%%%%%%%%%%%%%%%%%%%%%%

\parheader{Utility loss.}
We extract the base descriptors $\boldsymbol{d}_{\text{base}}$ from image patches $\boldsymbol{x}_\text{patch}$ and train NinjaNet ($\Theta$) with the descriptor learning loss from ~\cite{tian2019sosnet} to optimize NinjaDesc ($\boldsymbol{d}_{\text{ninja}}$).
\begin{equation}
    \mathcal{L}_{util}(\boldsymbol{x}_{\text{patch}}; \Theta) = \mathcal{L}_{triplet}(\boldsymbol{d}_{\text{ninja}}) + \mathcal{L}_{reg.}(\boldsymbol{d}_{\text{ninja}}),
    \label{eq:L_util}
\end{equation}
where $\mathcal{L}_{reg.}(\cdot)$ is the second-order similarity regularization term~\cite{tian2019sosnet}. 
We always freeze the weights of the base descriptor network, including the joint training process in Sec.~\ref{subsec:joint_adversarial_training}.
% This is also true for the joint training in Sec.~\ref{subsec:joint_adversarial_training}, 
%\ie only NinjaNet is being trained.
%\ie also for the joint training in Sec.~\ref{subsec:joint_adversarial_training}.
%Hence, we only achieve the initialization of NinjaNet's utility at this stage of training.

\subsection{Descriptor inversion model}
\label{subsec:inversion_model}

For our proposed adversarial learning framework, we utilize a descriptor inversion network as the adversary to reconstruct the input images from our NinjaDesc.
We adopt the UNet-based~\cite{unet} inversion network from prior work~\cite{invsfm,dangwal21}.
%
%\parheader{Architecture.}
Following Dangwal~\etal~\cite{dangwal21}, the inversion model $\Phi$ takes as input the sparse feature map $\textbf{F}_{\Theta} \in \mathbb{R}^{H \times W \times C}$ composed from the descriptors and their keypoints, and predicts the RGB image $\hat{\textbf{I}} \in \mathbb{R}^{h \times w \times 3}$, \ie $\hat{\textbf{I}} = \Phi(\textbf{F}_{\Theta})$. We denote $(H, W), (h, w)$ as the resolutions of the sparse feature image and the reconstructed RGB image, respectively.
$C$ is the dimensionality of the descriptor. %, \eg for SIFT~\cite{lowe2004sift}, SOSNet~\cite{tian2019sosnet} and NinjaDesc $C=128$.
%We use U-Net~\cite{unet} as $\Phi$, which has also been adopted in \cite{dangwal21} 
The detailed architecture is provided in the supplementary.
% Our input data is in the form of keypoint locations and descriptors, unlike the re-projections from the point-cloud in the earlier inversion model with VisibNet~\cite{pittaluga2019learning}, therefore as in \cite{dangwal21} we do not use VisibNet.
% 
% Furthermore, we also do not use the discriminator as in \cite{dangwal21} and its associated loss, since the training of the inversion model with the discriminator converges only after substantial amount of pre-training, and it improves the inversion model only very slightly.
% Hence, the discriminator does not affect the outcome of the joint training in Section~\ref{subsec:joint_adversarial_training} significantly.

\parheader{Reconstruction loss.}
The descriptor inversion model $\Phi$ is optimized under a reconstruction loss which is composed of two parts. 
The first loss is the mean absolute error (MAE) between the predicted $\hat{\textbf{I}}$ and input $\textbf{I}$ images,
\begin{equation}
    \mathcal{L}_{mae} = \sum_{i}^{h}\sum_{j}^{w} ||\hat{\textbf{I}}_{i,j} - \textbf{I}_{i,j}||_{1}.
    \label{eq:L_mae}
\end{equation}
The second loss is the perceptual loss, which is the L2 distance between intermediate features of a VGG16~\cite{simonyan2015VGG} network pretrained on ImageNet~\cite{deng2009ImageNet},
\begin{equation}
    \mathcal{L}_{perc} = \sum_{k=1}^{3} \sum_{i}^{h_k}\sum_{j}^{w_k} || \textbf{$\psi$}_{k,i,j}^{VGG}(\hat{\textbf{I}}) - \textbf{$\psi$}_{k,i,j}^{VGG}(\textbf{I}) ||_{2}^{2},
    \label{eq:L_perc}
\end{equation}
where $\textbf{$\psi$}_k^{VGG}(\textbf{I})$ are the feature maps extracted at layers $k \in \{ 2,9,16 \}$, and $(h_{k}, w_{k})$ is the corresponding resolution.

The reconstruction loss is the sum of the two terms
\begin{equation}
    \mathcal{L}_{recon} (\boldsymbol{x}_\text{image}; \Phi) = \mathcal{L}_{mae} + \mathcal{L}_{perc}.
    \label{eq:L_recon}
\end{equation}
where $\boldsymbol{x}_{\text{image}}$ denote the image data term that includes both the descriptor feature map $\textbf{F}_{\Theta}$ and the RGB image $\textbf{I}$. 

\parheader{Reconstruction initialization.}
For the joint adversarial training described in Sec.~\ref{subsec:joint_adversarial_training}, we initialize the the inversion model using the initialized NinjaDesc in Sec.~\ref{subsec:privacy_encoder}, 
This part is done using the MegaDepth~\cite{li2018megadepth} dataset, which contains images of landmarks across the world. For the keypoint detection we use the Harris corners~\cite{harris_corner} in our experiments. %and crop patches around each of them to form the sparse feature image as an input to the network.

\subsection{Joint adversarial training}
\label{subsec:joint_adversarial_training}
The central component of engineering our \privacying NinjaDesc is the joint adversarial training step, which is illustrated in Fig.~\ref{fig:joint_training} and elaborated as pseudo-code in Algorithm~\ref{alg:joint_training}.
We aim to minimize trade-off between utility and privacy, which are the two competing objectives.
Inspired by methods using adversarial learning~\cite{goodfellow2014generative,xie2017controllable, roy2019mitigating}, we formulate the optimization of utility and privacy trade-off as an adversarial learning process.
%%%%%%%%%%%%%%%%%%%%%%%%%%%%%%%%%%%%%%%%%%%%%%%%%%%%%%%%%%%%%%%
%%%%%%%%%%%%%%%%%%%%%%%%%%%%%%%%%%%%%%%%%%%%%%%%%%%%%%%%%%%%%%%
\setlength{\textfloatsep}{8pt}% Remove \textfloatsep
\begin{algorithm}[t!]
\caption{Pseudo-code for the joint adversarial training process of NinjaDesc}\label{alg:joint_training}
    \begin{algorithmic}[1]
    \State NinjaNet: $\Theta_0$ $\leftarrow$ initialize with Eqn.~\ref{eq:L_util}
    \State Desc. inversion model: $\Phi_0$ $\leftarrow$ initialize with Eqn.~\ref{eq:L_recon}
    \State $\lambda$ $\leftarrow$ set \losslambda
        \For{$i \leftarrow 1, \text{number of iterations}$}
           \If {$i = 0$}
           \State $\Theta \leftarrow \Theta_0, \Phi \leftarrow \Phi_0$ 
           \EndIf
           \State Compute $\mathcal{L}_{util}$ from $\boldsymbol{x}_{\text{patch}}$ and $\Theta$.
           \State Extract sparse features on $\boldsymbol{x}_{\text{image}}$ with $\Theta$,
           
           reconstruct image with $\Phi$
           
           and compute 
           $\mathcal {L}_{recon}\left(\boldsymbol{x}_{\text{image}}; \Theta, \Phi \right)$.
           \State Update weights of $\Theta$:
           \vspace{-8pt}
           \begin{equation*}
               \Theta^{'} \leftarrow \nabla_\Theta \left(\mathcal{L}_{util} - \lambda \mathcal{L}_{recon} \right).
           \vspace{-4pt}
           \end{equation*}
           \State Extract sparse features on $\boldsymbol{x}_{\text{image}}$ with $\Theta^{'}$,
           
           reconstruct image with $\Phi$
           
           and
           compute $\mathcal {L}_{recon}(\boldsymbol{x}_{\text{image}}; \Theta^{'}, \Phi )$.
           \State Update weights of $\Phi$:
           \vspace{-8pt}
           \begin{equation*}
               \Phi^{'} \leftarrow \nabla_\Phi \mathcal{L}_{util}.
       \vspace{-4pt}
           \end{equation*}
           \State $\Theta \leftarrow \Theta^{'}, \Phi \leftarrow \Phi^{'}$ 
        \EndFor
        \end{algorithmic}
\end{algorithm}

%%%%%%%%%%%%%%%%%%%%%%%%%%%%%%%%%%%%%%%%%%%%%%%%%%%%%%%%%%%%%%%
%%%%%%%%%%%%%%%%%%%%%%%%%%%%%%%%%%%%%%%%%%%%%%%%%%%%%%%%%%%%%%%
%The objective of our NinajaNet $\Theta$ is to fool conceals the visual content by maximizing this error. 
The objective of the descriptor inversion model $\Phi$ is to minimize the reconstruction error over image data $\boldsymbol{x}_{\text{image}}$. 
On the other hand, NinajaNet $\Theta$ aims to conceal the visual content by maximizing this error. 
Thus, the resulting objective function for content concealment $V(\Theta, \Phi)$ is a minimax game between the two:
\begin{equation}
    \min_{\Phi}\max_{\Theta} V(\Theta, \Phi) =  \mathcal{L}_{recon}\left(\boldsymbol{x}_{\text{image}}; \Theta, \Phi\right).
    \label{eq:minimax}
\end{equation}
At the same time, we wish to maintain the descriptor utility:
\begin{equation}
    \min_{\Theta} \mathcal{L}_{util}(\boldsymbol{x}_{\text{patch}}; \Theta).
    \label{eq:max_util}
\end{equation}
%----------------------------------------------------------------------------------
%%%%%%%%%%%%%%%%%%%%%%%%%%%%%%%%%%%%%%%%%%%%%%%%%%%%%%%%%%%%%%%%%%%%%%%%%%%%%%%%%%%
%%%%%%%%%%%%%%%%%%%%%%%%%%%%%%%%%%%%%%%%%%%%%%%%%%%%%%%%%%%%%%%%%%%%%%%%%%%%%%%%%%%
\begin{figure*}[!ht]
    \vspace{-0pt}
    \centering
    \includegraphics[width=\linewidth, trim={
        0pt, 0pt, 0pt, 0pt}, clip]{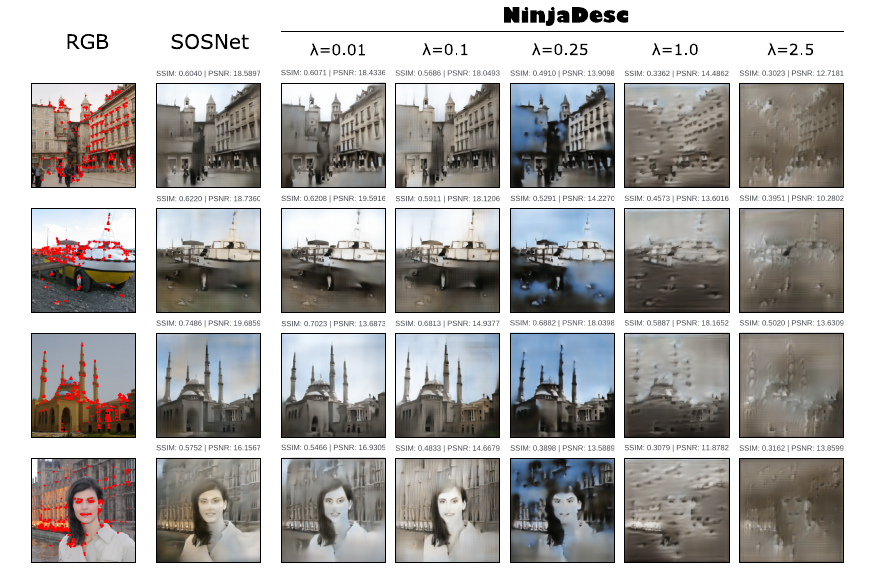}
    \vspace{-23pt}
    \caption{
        Qualitative results on landmark images. First column: original images overlaid with the 1000 (red) Harris corners~\cite{harris_corner}. Second column: reconstructions by the inversion model from raw SOSNet~\cite{tian2019sosnet} descriptors extracted on those points. The last five columns show reconstruction from NinjaDesc with increasing \losslambda $\lambda$. The SSIM and PSNR \wrt the original images are shown on top of each reconstruction. Best viewed digitally. \textit{
        Image credits: first 3 --- Holidays dataset~\cite{jegou2008hamming}; last --- \href{https://www.flickr.com/photos/laylamoran4battersea/4543311974/}{\textit{laylamoran4battersea}} (Flickr).
    }
    }
    \label{fig:recon_landmakrs}
    \vspace{-10pt}
\end{figure*}%%
%%%%%%%%%%%%%%%%%%%%%%%%%%%%%%%%%%%%%%%%%%%%%%%%%%%%%%%%%%%%%%%%%%%%%%%%%%%%%%%%%%%
This brings us to the two separate optimization objectives for $\Theta$ and $\Phi$ that we will describe in the following.
For the inversion model $\Phi$, the objective remains the same as in Eqn.~\ref{eq:minimax}:
\begin{equation} \mathcal{L}_\Phi = \mathcal{L}_{recon}\left(\boldsymbol{x}_{\text{image}} ; \Theta, \Phi\right).
\label{eq:L_Phi} 
\end{equation} 
%  \begin{equation}
%      \mathcal{L}_\Phi = \mathcal{L}_{recon}\left(\boldsymbol{x}_{\text{image}} ; \Theta, \Phi\right).
%      \label{eq:L_Phi}
%  \end{equation}
%
However, for maintaining utility, NinjaNet with weights $\Theta$ is also optimized with the utility loss $\mathcal{L}_{util}(\boldsymbol{x}_{\text{patch}}; \Theta)$ from Eqn.~\ref{eq:L_util}.
%used for utility initialization with loss $\mathcal{L}_{util}$ from Eqn.~\ref{eq:L_util}.
In conjunction with the maximization by $\Theta$ from Eqn.~\ref{eq:minimax}, the loss for NinjaNet becomes
\begin{equation}
    \mathcal{L}_\Theta = \mathcal{L}_{util}\left(\boldsymbol{x}_{\text{patch}}; \Theta\right) - \lambda \mathcal{L}_{recon}\left(\boldsymbol{x}_{\text{image}}; \Theta, \Phi\right),
    \label{eq:L_Theta}
\end{equation}
where $\lambda$ controls the balance of how much $\Theta$ prioritizes \privacy over utility retention, \ie the \losslambda.
In practice, we optimize $\Theta$ and $\Phi$ in an alternating manner, such that $\Theta$ is not optimized in Eqn.~\ref{eq:L_Phi} and $\Phi$ is not optimized in Eqn.~\ref{eq:L_Theta}.
%Note that although the gradients from $\Phi$ are still required for the reconstruction loss, $\Phi$ is not optimized in Eqn.~\ref{eq:L_Theta}.
The overall objective is then 
\begin{equation}
    \Theta^*, \Phi^* = \operatorname*{\arg\min}_{\Theta, \Phi} (\mathcal{L}_{\Theta} + \mathcal{L}_{\Phi}).
\end{equation}
% where the subscript denotes which network is optimized \wrt to each loss function.

%-------------------------------------------------------------------
\subsection{Implementation details}
\label{subsec:implementation_details}
The code is implemented using PyTorch~\cite{paszke2019PyTorch}. We use Kornia~\cite{riba2020kornia}'s implementation of SIFT for GPU acceleration. For all training, we use the Adam~\cite{adam_optimizer} optimizer with $(\beta_1, \beta_2) = (0.9, 0.999)$ and $\lambda = 0$.

\parheader{Utility initialization.} We use the \textit{liberty} set of the UBC patches~\cite{goesele2007ubc} to train NinjaNet for 200 epochs and select the model with the lowest average FPR@95 in the other two sets (\textit{notredame} and \textit{yosemite}).
The number of submodules in NinjaNet ($N$ in Fig.~\ref{fig:encoder}) is $N=1$, since we observed no improvement in FPR@95 by increasing $N$. % (\ie the complexity).
Dropout rate is 0.1.
We use a batch-size of 1024 and learning rate of 0.01.

\parheader{Reconstruction initialization.}
We randomly split MegaDepth~\cite{li2018megadepth} into train\,/\,validation\,/\,test split of ratio 0.6\,/\,0.1\,/\,0.3. % and train the inversion model on the train split.
The process of forming a feature map is the same as in \cite{dangwal21} and we use up to 1000 Harris corners~\cite{harris_corner} for all experiments.
We train the inversion model with a batch-size of 64, learning rate of 1e-4 for a maximum of 200 epochs and select the best model with the lowest structural similarity (SSIM) on the validation split. 
%Like \cite{dangwal21}, we do not use VisibNet from \cite{invsfm} as our input is not of the form of point-cloud reprojections.
% 
We also do not use the discriminator as in \cite{dangwal21}, since convergence of the discriminator takes substantially longer, and it improves the inversion model only very slightly.
%the training of the inversion model with the discriminator converges only after substantial amount of pre-training, and it improves the inversion model only very slightly, see Table~\ref{tab:recon_results}.
% Hence, the discriminator does not affect the outcome of the joint training in Section~\ref{subsec:joint_adversarial_training} significantly.

\parheader{Joint adversarial training.} The dataset configurations for $\mathcal{L}_{util}$ and $\mathcal{L}_{recon}$ are the same as in the above two steps, except the batch size, that is 968 for UBC patches.
We use equal learning rate for $\Theta$ and $\Phi$. This is 5e-5 for SOSNet~\cite{tian2019sosnet} and HardNet~\cite{mishchuk2017hardnet}, and 1e-5 for SIFT~\cite{lowe2004sift}.
NinjaDesc with the best FPR@95 in 20 epochs on the validation set is selected for testing.

%We train with 8 Tesla V100 GPUs and 256GB of RAM. The three training steps take roughly 1 hour, 4 days and 12 hours respectively.

\setlength{\textfloatsep}{20pt}
% The implementation details of all three stages of training are in the Supplementary Material (S.M.).
\vspace{-0pt}
\section{Experimental results}
\label{sec:results}
In this section, we evaluate NinjaDesc on the two criteria that guide its design --- the ability to simultaneously achieve: (1) content concealment (privacy) and (2) utility (matching accuracy and camera localization performance).
\subsection{Content concealment (Privacy)}
\label{subsec:results_privacy}
We assess the content-concealing ability of NinjaDesc by measuring the reconstruction quality of descriptor inversion attacks.
Here we assume the inversion model has access to the NinjaDescs and the input RGB images for training, \ie $\boldsymbol{x}_{\text{image}}$ in Sec.~\ref{subsec:inversion_model}.
We train the inversion model from scratch for NinjaDesc (Eqn.~\ref{eq:L_recon}) on the train split of MegaDepth~\cite{li2018megadepth}, and the best model with the highest SSIM on the validation split is used for the evaluation.

%%%%%%%%%%%%%%%%%%%%%%%%%%%%%%%%%%%%%%%%%%%%%%%%%%%%%%%%%%%%%%%%%%%%%%%
%%%%%%%%%%%%%%%%%%%%%%%%%%%%%%%%%%%%%%%%%%%%%%%%%%%%%%%%%%%%%%%%%%%%%%%
\begin{figure*}[!t]
    \vspace{-19pt}
    \centering
    \includegraphics[width=\linewidth, trim={
        -10pt, 5pt, 15pt, 30pt}, clip]{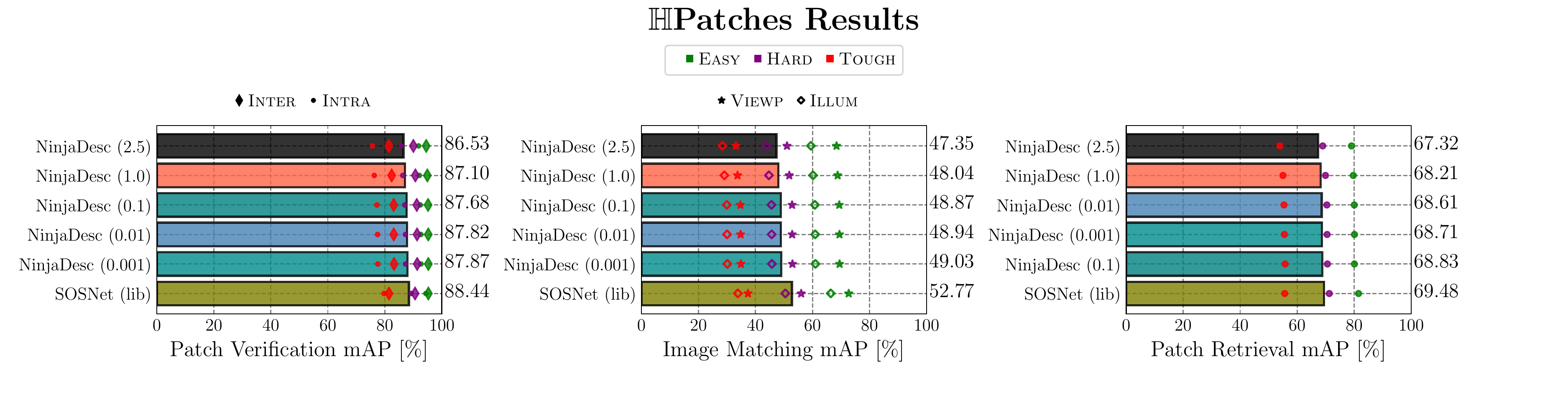}
    \vspace{-28pt}
    \caption{HPatches evaluation results. We compare the baseline SOSNet~\cite{tian2019sosnet} \vs NinjaDesc, with 5 different levels of \losslambda $\lambda$ (indicated by the number in parenthesis). All results are from models trained on the \textit{liberty} subset of the UBC patches~\cite{goesele2007ubc} dataset.}
    \label{fig:hpatches_results}
    \vspace{-10pt}
\end{figure*}

%%%%%%%%%%%%%%%%%%%%%%%%%%%%%%%%%%%%%%%%%%%%%%%%%%%%%%%%%%%%%%%%%%%%%%%
%%%%%%%%%%%%%%%%%%%%%%%%%%%%%%%%%%%%%%%%%%%%%%%%%%%%%%%%%%%%%%%%%%%%%%%

Recall in Eqn.~\ref{eq:L_Theta}, $\lambda$ is the \losslambda controlling how much NinjaDesc prioritizes privacy over utility.
The intuition is that, the higher $\lambda$ is, the more aggressive NinjaDesc tries to prevent reconstruction quality by the inversion model.
We perform descriptor inversion on NinjaDesc that are trained with a range of $\lambda$ values to demonstrate its effect on reconstruction quality. % both qualitatively and quantitatively.
%The training is done on the train split of MegaDepth~\cite{li2018megadepth}, and the best model with the highest SSIM on the validation split is used for the evaluation. 
%For a given type of descriptors, we train a descriptor inversion model from scratch (Eqn.~\ref{eq:L_recon}) on the train split of MegaDepth~\cite{li2018megadepth}, and the best model with the highest SSIM on the validation split is used for the evaluation.

%%%%%%%%%%%%%%%%%%%%%%%%%%%%%%%%%%%%%%%%%%%%%%%%%%%%%%%%%%%%%%%%%%%%%%%%%%%%%%%%%%%%%%%
%%%%%%%%%%%%%%%%%%%%%%%%%%%%%%%%%%%%%%%%%%%%%%%%%%%%%%%%%%%%%%%%%%%%%%%%%%%%%%%%%%%%%%%
\begin{table}[t]
\setlength{\tabcolsep}{2.3pt}
\renewcommand{\arraystretch}{1.05}
	\footnotesize
	\centering
	\begin{tabular}{r | c c c c c c c c c}
		\toprule
		 & & \multirowcell{2.8}{SOSNet \\ (Raw)} & \multicolumn{6}{c}{NinjaDesc ($\lambda$)}  \\
		\cmidrule{5-10}
		Metric\,  &   & &  & 0.001  & 0.01   & 0.1    & 0.25   & 1.0    & 2.5         \\
		\midrule
		MAE ($\uparrow$)\,   &    & 0.104  && 0.117  & 0.125  & 0.129  & 0.162  & 0.183  & 0.212   \\
		SSIM ($\downarrow$)\, &     & 0.596  && 0.566  & 0.569  & 0.527  & 0.484  & 0.385  & 0.349   \\
		PSNR ($\downarrow$)\,  &     & 17.904 && 18.037 & 16.826 & 17.821 & 17.671 & 13.367 & 12.010 \\
	
		\bottomrule
	\end{tabular}
	\vspace{-6pt}
	\caption{Quantitative results of the descriptor inversion on SOSNet \vs NinjaDesc, evaluated on the MegaDepth~\cite{li2018megadepth} test split\protect\footnotemark. The arrows indicate higher\,/\,lower value is better for privacy.}
	\label{tab:recon_results}
	\vspace{-12pt}
\end{table}

%%%%%%%%%%%%%%%%%%%%%%%%%%%%%%%%%%%%%%%%%%%%%%%%%%%%%%%%%%%%%%%%%%%%%%%%%%%%%%%%%%%%%%%
%%%%%%%%%%%%%%%%%%%%%%%%%%%%%%%%%%%%%%%%%%%%%%%%%%%%%%%%%%%%%%%%%%%%%%%%%%%%%%%%%%%%%%%

Fig.~\ref{fig:recon_landmakrs} shows qualitative results of descriptor inversion attacks when changing $\lambda$.
We observe that $\lambda$ indeed fulfills the role of controlling how much NinjaDesc conceals the original image content. 
When $\lambda$ is small, \eg $0.01, 0.1$, the reconstruction is only slightly worse than that from the baseline SOSNet. 
As $\lambda$ increases to $0.25$, there is a visible deterioration in quality.
Once equal\,/\,stronger weighting is given to privacy ($\lambda=1, 2.5$), little texture\,/\,structure is revealed, achieving high privacy.

Such observation is also validated quantitatively by Table~\ref{tab:recon_results}, where we see a drop in performance of the inversion model as $\lambda$ increases across the three metrics: average mean absolute error (MAE), structural similarity (SSIM), and peak signal-to-noise ratio (PSNR) which are computed from the reconstructed image and the original input image.
%%%%%%%%%%%%%%%%%%%%%%%%%%%%%%%%%%%%%%%%%%%%%%%%%%%%%%%%%%%%%%%%%%%%%%%%%%%%%%%%%%%%%%%%%%%%%%%%%%%
\footnotetext{Note that in \cite{dangwal21}, only SSIM is reported, and we do not share the same train\,/\,validation\,/\,test split. Also, \cite{dangwal21} uses the discriminator loss for training which we omit, and it leads to slight difference in SSIM.}
%%%%%%%%%%%%%%%%%%%%%%%%%%%%%%%%%%%%%%%%%%%%%%%%%%%%%%%%%%%%%%%%%%%%%%%%%%%%%%%%%%%%%%%%%%%%%%%%%%%

\subsection{Utility retention}
\label{subsec:results_utility}
We measure the utility of NinjaDesc via two tasks: image matching and visual localization.

\parheader{Image matching.} We evaluate NinjaDesc based on SOSNet~\cite{tian2019sosnet} with a set of different \losslambda on the HPatches~\cite{balntas2017hpatches} benchmarks, which is shown in Fig.~\ref{fig:hpatches_results}. 
NinjaDesc is comparable with SOSNet in mAP across all three tasks, especially for the \textit{verification} and \textit{retrieval} tasks.
Also, higher \losslambda $\lambda$ generally corresponds to lower mAP, as $\mathcal{L}_{util}$ becomes less dominant in Eqn.~\ref{eq:L_Theta}.

%%%%%%%%%%%%%%%%%%%%%%%%%%%%%%%%%%%%%%%%%%%%%%%%%%%%%%%%%%%%%%%%%%%%%%%%%%%%%%%%%%%%%%%%
%%%%%%%%%%%%%%%%%%%%%%%%%%%%%%%%%%%%%%%%%%%%%%%%%%%%%%%%%%%%%%%%%%%%%%%%%%%%%%%%%%%%%%%%
\begin{figure*}[!ht]
    \vspace{-0pt}
    \centering
    \includegraphics[width=.95\linewidth, trim={
        0pt, 0pt, 0pt, 0pt}, clip]{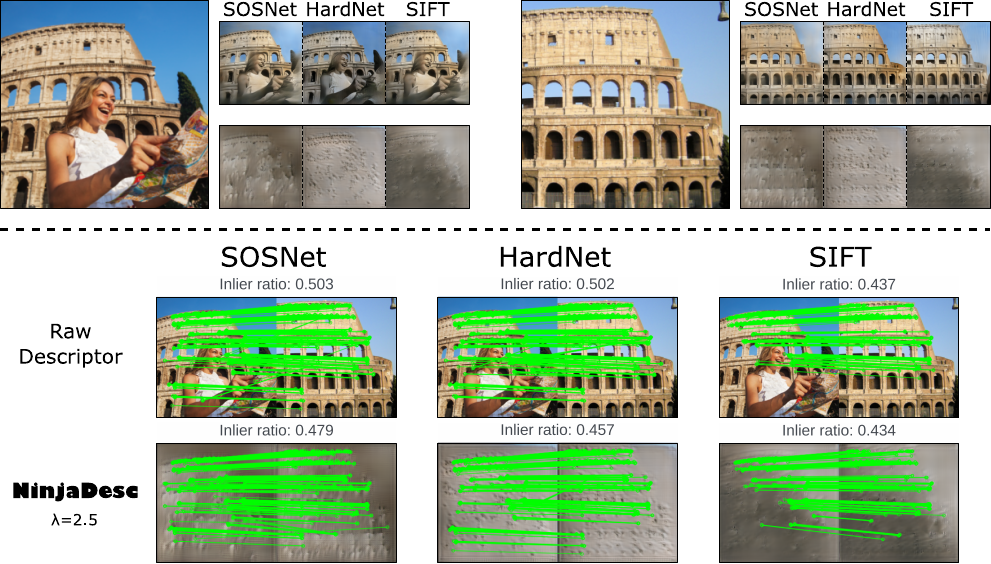}
    \vspace{-8pt}
    %\vspace{-19pt}
    \caption{Illustration of our proposed adversarial descriptor learning framework's generalization across three different base descriptors.
        \textbf{Top.} We show two matching images. 
        Two rows of small images to the right of each of them are the reconstructions. 
        The top \& bottom rows are, respectively, the reconstructions from the raw descriptor and from NinjaDesc ($\lambda=2.5$) associated with the base descriptor above.
        \textbf{Bottom.} We visualize the matches between the two images on raw descriptors \vs NinjaDesc ($\lambda=2.5$) for each of the three base descriptors.
        \textit{Image credits: left --- \href{https://stock.adobe.com/images/woman-on-the-background-of-the-colosseum/200999902?prev_url=detail}{Tatyana Gladskih}\,/\,stock.adobe.com; right --- \href{https://commons.wikimedia.org/wiki/File:Colosseum_Roma_2009.jpg}{Urse Ovidiu} (Wikimedia Commons, Public Domain)}.
        }
    \label{fig:recon_landmark_all_descs}
    \vspace{-10pt}
\end{figure*}

%%%%%%%%%%%%%%%%%%%%%%%%%%%%%%%%%%%%%%%%%%%%%%%%%%%%%%%%%%%%%%%%%%%%%%%%%%%%%%%%%%%%%%%%
%%%%%%%%%%%%%%%%%%%%%%%%%%%%%%%%%%%%%%%%%%%%%%%%%%%%%%%%%%%%%%%%%%%%%%%%%%%%%%%%%%%%%%%%

\begin{table}[t!]
\setlength{\tabcolsep}{2.0pt}
\renewcommand{\arraystretch}{1.00}
	% \footnotesize
	\scriptsize
	\centering
	\resizebox{.48\textwidth}{!}{
	    \begin{tabular}{c c c c c c}
		\toprule
		& & \multirow{3}{*}{Method} & \multicolumn{3}{c}{Accuracy @ Thresholds (\%)} \\
		\multirow{2}{*}{Query} & \multirow{2}{*}{NNs} & & $0.25$m, $2^{\circ}$ & $0.5$m, $5^{\circ}$ & $5.0$m, $10^{\circ}$ \\ 
		\cmidrule{4-6}
		& & \tiny{Base Desc} &  \tiny{~SOS~/~Hard~/~SIFT~}  & \tiny{~SOS~/~Hard~/~SIFT~} & \tiny{~SOS~/~Hard~/~SIFT~}\\
		\midrule
		\multirowcell{8}{Day\\($824$)} & \multirowcell{4.5}{20} 
    	& Raw               & ~85.1\,/\,85.4\,/\,84.3~ & ~92.7\,/\,93.1\,/\,92.7~ & ~97.3\,/\,98.2\,/\,97.6~ \\
		& & $\lambda=0.1$   & ~85.4\,/\,84.7\,/\,82.0~ & ~92.5\,/\,91.9\,/\,91.1~ & ~97.5\,/\,96.8\,/\,96.4~ \\
		& & $\lambda=1.0$   & ~84.7\,/\,84.3\,/\,82.9~ & ~92.4\,/\,91.9\,/\,91.0~ & ~97.2\,/\,96.7\,/\,96.1~ \\
		& & $\lambda=2.5$   & ~84.6\,/\,83.7\,/\,82.5~ & ~92.4\,/\,92.0\,/\,91.0~ & ~97.1\,/\,96.8\,/\,96.0~ \\
		\cmidrule{3-6}
		& \multirowcell{4.5}{50} 
		& Raw               & ~85.9\,/\,86.8\,/\,86.0~ & ~92.5\,/\,93.7\,/\,94.1~ & ~97.3\,/\,98.1\,/\,98.2~ \\
		& & $\lambda=0.1$   & ~85.2\,/\,85.2\,/\,84.2~ & ~92.2\,/\,92.4\,/\,91.4~ & ~97.1\,/\,97.1\,/\,96.6~ \\
		& & $\lambda=1.0$   & ~84.7\,/\,85.7\,/\,83.4~ & ~92.2\,/\,92.6\,/\,91.6~ & ~97.2\,/\,96.7\,/\,96.7~ \\
		& & $\lambda=2.5$   & ~85.6\,/\,85.3\,/\,83.6~ & ~92.7\,/\,91.7\,/\,91.1~ & ~97.3\,/\,96.8\,/\,96.2~ \\
		\cmidrule{2-6}
		\multirowcell{5.5}{Night\\($191$)} & \multirowcell{4.5}{20} & 
		Raw               & ~49.2\,/\,52.4\,/\,50.8~ & ~60.2\,/\,62.3\,/\,62.3~ & ~68.1\,/\,72.3\,/\,72.8~ \\
		& & $\lambda=0.1$ & ~47.6\,/\,43.5\,/\,44.0~ & ~57.1\,/\,54.5\,/\,51.3~ & ~63.4\,/\,61.8\,/\,61.3~ \\
		& & $\lambda=1.0$ & ~45.5\,/\,44.5\,/\,41.4~ & ~56.0\,/\,51.8\,/\,52.9~ & ~61.8\,/\,60.2\,/\,62.3~ \\
		& & $\lambda=2.5$ & ~45.0\,/\,44.5\,/\,43.5~ & ~55.0\,/\,54.5\,/\,49.7~ & ~61.8\,/\,61.3\,/\,61.3~ \\
		\cmidrule{3-6}
		  & \multirowcell{4.5}{50} & 
		Raw                 & ~44.5\,/\,47.6\,/\,51.3~ & ~52.4\,/\,59.7\,/\,62.3~ & ~60.2\,/\,64.9\,/\,74.3~ \\
		& & $\lambda=0.1$   & ~39.8\,/\,39.8\,/\,41.9~ & ~47.6\,/\,48.7\,/\,50.3~ & ~57.6\,/\,56.0\,/\,59.7~ \\
		& &  $\lambda=1.0$  & ~42.9\,/\,39.8\,/\,39.8~ & ~52.4\,/\,49.2\,/\,48.2~ & ~57.1\,/\,54.5\,/\,56.5~ \\
		& & $\lambda=2.5$   & ~41.9\,/\,38.2\,/\,40.3~ & ~49.2\,/\,47.1\,/\,49.2~ & ~56.6\,/\,55.0\,/\,57.1~ \\
		
		\bottomrule
	\end{tabular}
	}
	\vspace{-6.0pt}
	\caption{Visual localization results on Aachen-Day-Night v1.1~\cite{zhang2020aachenv_1_1}. `Raw' corresponds to the base descriptor in each column, followed by three $\lambda$ vales (0.1, 1.0, 2.5) for NinjaDesc.}
	\label{tab:aachen}
	\vspace{-10pt}
\end{table}

\parheader{Visual localization.} We evaluate NinjaDesc with three base descriptors - SOSNet~\cite{tian2019sosnet}, HardNet~\cite{mishchuk2017hardnet} and SIFT~\cite{lowe2004sift} on the Aachen-Day-Night v1.1~\cite{sattler2018benchmarking,zhang2020aachenv_1_1} dataset using the Kapture~\cite{kapture2020} pipeline.
We use AP-Gem~\cite{revaud2019aploss} for retrieval and localize with the shortlist size of 20 and 50. The keypoint detector used is DoG~\cite{lowe2004sift}.
Table~\ref{tab:aachen} shows localization results. 
Again, we observe little drop in accuracy for NinjaDesc overall compared to the original base descriptors, ranging from low ($\lambda=0.1$) to high ($\lambda=2.5$) privacies.
Comparing our results on HardNet and SIFT with Table 3 in Dusmanu \etal~\cite{dusmanu2020privacy}, NinjaDesc is noticeably better in retaining the visual localization accuracy of the base descriptors than the subspace descriptors in \cite{dusmanu2020privacy}\footnote{
\cite{dusmanu2020privacy} is evaluated on Aachen-Day-Night v1.0, resulting in higher accuracy in \textit{Night} due to poor ground-truths, and the code of~\cite{dusmanu2020privacy} is not released yet. 
We also report our results on v1.0 in the supplementary. % for better comparison.
},
\eg drop in \textit{night} is up to 30\% for HardNet in \cite{dusmanu2020privacy} but $\approx10\%$ for NinjaDesc.

Hence, the results on both image matching and visual localization tasks demonstrate that NinjaDesc is able to retain the majority of its utility \wrt to the base descriptors.

\vspace{-0pt}
\section{Ablation studies}
\label{sec:ablation_studies}
Table~\ref{tab:aachen} already hints that our proposed adversarial descriptor learning framework generalizes to several base descriptors in terms of retaining utility. % for visual localization. 
In this section, we further investigate the generalizability of our method through additional experiments on different types of descriptors, inversion network architectures, and scene categories.  %, as well as conducting an analysis on the trade-off between utility and privacy in NinjaDesc. %,show that it also generalizes to different descriptors and architecture for \privacy.
%We conduct an analysis on the trade-off between utility and privacy in NinjaDesc.
%Lastly, we show that it generalizes to faces even though it is trained on landmark images.

\vspace{-0pt}
\subsection{Generalization to different descriptors}
\label{subsec:generalization_descs}
%%%%%%%%%%%%%%%%%%%%%%%%%%%%%%%%%%%%%%%%%%%%%%%%%%%%%%%%%%%%%%%%%%%%%%%%%%%%%%%%%%%%%%%%
%%%%%%%%%%%%%%%%%%%%%%%%%%%%%%%%%%%%%%%%%%%%%%%%%%%%%%%%%%%%%%%%%%%%%%%%%%%%%%%%%%%%%%%%
\begin{table}[t!]
\setlength{\tabcolsep}{3pt}
\setlength{\aboverulesep}{0pt}
\setlength{\belowrulesep}{0pt}
\renewcommand{\arraystretch}{1.1}
	\footnotesize
	\centering
	\begin{tabular}{r | c c c c c c}
		\toprule 
		 & \multicolumn{6}{c}{SSIM ($\downarrow$)} \\
		%\midrule
		%\midrule
		\cmidrule{2-7}
		\multirow{2}{*}{Base Descriptor} & \multirowcell{2}{Raw \\ (w/o NinjaDesc)} & \multicolumn{5}{c}{NinjaDesc ($\lambda$)} \\
		\cmidrule{3-7}
		 & &  0.01  & 0.1 &  0.25  &  1.0    &  2.5   \\
		\midrule
		SOSNet  &  0.596  & 0.569  &  0.527  &  0.484  & 0.385  &  0.349 \\
		HardNet &  0.582  & 0.545 & 0.516 & 0.399 & 0.349  & 0.312 \\
		SIFT    &  0.553  & 0.490  &  0.459  &  0.395  & 0.362  &  0.296 \\
		\bottomrule
	\end{tabular}
	\vspace{-8pt}
	\caption{Qualitative performance of the descriptor inversion model on the MegaDepth~\cite{li2018megadepth} test split with three base descriptors and the corresponding NinjaDescs, varying in \losslambda.}
	\label{tab:recon_all_descs}
	\vspace{-10pt}
\end{table}
%%%%%%%%%%%%%%%%%%%%%%%%%%%%%%%%%%%%%%%%%%%%%%%%%%%%%%%%%%%%%%%%%%%%%%%%%%%%%%%%%%%%%%%%
%%%%%%%%%%%%%%%%%%%%%%%%%%%%%%%%%%%%%%%%%%%%%%%%%%%%%%%%%%%%%%%%%%%%%%%%%%%%%%%%%%%%%%%%
We extend the same experiments from SOSNet~\cite{tian2019sosnet} in Table~\ref{tab:recon_results} to include HardNet~\cite{mishchuk2017hardnet} and SIFT~\cite{lowe2004sift} as well. We report SSIM in Table~\ref{tab:recon_all_descs}.
Similar to the observation for SOSNet, increasing \losslambda $\lambda$ reduces reconstruction quality for both HardNet and SIFT as well.
In Fig.~\ref{fig:recon_landmark_all_descs}, we qualitatively show the descriptor inversion and correspondence matching result across all three base descriptors. % on a pair of matching image.
We observe that NinjaDesc derived from all three base descriptors are effective in %Fig.~\ref{fig:recon_landmark_all_descs}, choosing a high \losslambda ($\lambda$ = 2.5) results in
concealing important contents such as person or landmark compared with the raw base descriptors.
The visualization of keypoint correspondences between the images also demonstrates the utility retention of our proposed learning framework across different base descriptors. 

\subsection{Generalization to different architectures}
\label{subsec:generalization_architecture}
\begin{table}[t]
\setlength{\tabcolsep}{3pt}
\renewcommand{\arraystretch}{1.1}
\setlength{\aboverulesep}{0pt}
\setlength{\belowrulesep}{0pt}
	\footnotesize
	\centering
	\begin{tabular}{r | c c c c  c c c}
		\toprule
		\scriptsize{Arch.} & \multicolumn{3}{c}{UNet} & & \multicolumn{3}{c}{UResNet} \\
		\cmidrule{2-4} \cmidrule{6-8}
        & \scriptsize{SOSNet}   & \scriptsize{$\lambda=1.0$} & \scriptsize{$\lambda=2.5$} & & \scriptsize{SOSNet}  & \scriptsize{$\lambda=1.0$}  & \scriptsize{$\lambda=2.5$}  \\
		\midrule
		MAE ($\uparrow$)\, & 0.104  & 0.183  & 0.212  & \,  & 0.121  & 0.190  & 0.202  \\
		SSIM ($\downarrow$)\,  & 0.596  & 0.385  & 0.349  & \,  & 0.595  & 0.427  & 0.380 \\
		PSNR ($\downarrow$)\,  & 17.904 & 13.367 & 12.010 & \,  & 16.533 & 12.753 & 12.299 \\
	
		\bottomrule
	\end{tabular}
	\vspace{-5pt}
	\caption{Reconstruction results on MegaDepth~\cite{li2018megadepth}. We compare the UNet used in this work \vs a different architecture --- UResNet.}
	\label{tab:recon_architecture}
	\vspace{-10pt}
\end{table}

So far, all experiments are evaluated with the same architecture for the inversion model - the UNet~\cite{unet}-based network~\cite{dangwal21,invsfm}.
To verify that NinjaDesc does not overfit to this specific architecture, we conduct a descriptor inversion attack using an inversion model with drastically different architecture, called UResNet, which has a ResNet50~\cite{he2016ResNet} as the encoder backbone and residual decoder blocks. (See the supplementary material.)
The results are shown in Table~\ref{tab:recon_architecture}, which depict only SSIM is slightly improved compared to UNet whereas MAE and PSNR remain relatively unaffected.
This result illustrate that our proposed method is not limited by the architectures of the inversion model. % the descriptor inversion model.
% \subsection{Number of keypoints}

\subsection{\Privacy on faces}
\label{subsec:faces}
\begin{figure}[!t]
    \vspace{-0pt}
    \centering
    \includegraphics[width=\linewidth, trim={
        5pt, 0pt, 5pt, 5pt}, clip]{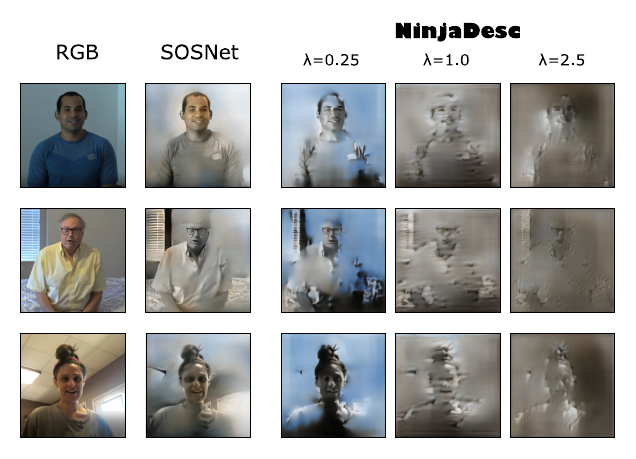}
    \vspace{-22pt}
    \caption{Qualitative reconstruction results on faces. Images are cropped frames sampled from videos in the DFDC~\cite{DFDC} dataset.}
    \label{fig:recon_faces}
    \vspace{-2pt}
\end{figure}

%Apart from landmarks images which we evaluate (and train) NinajaDesc on so far, Fig.~\ref{fig:recon_faces} 
We further show qualitative results on human faces using the Deepfake Detection Challenge (DFDC)~\cite{DFDC} dataset. Fig.~\ref{fig:recon_faces} presents the descriptor inversion result using the base descriptors (SOSNet~\cite{tian2019sosnet}) as well as our NinjaDesc varying in \losslambda $\lambda$.
Similar to what we observed in Fig.~\ref{fig:recon_landmakrs}, we see progressing concealment of facial features as we increase $\lambda$ compared to the reconstruction on SOSNet. 

\vspace{-2ex}
\section{Utility and privacy trade-off}
\label{sec:trade-off}
We now describe two experiments we perform to further investigate the utility and privacy trade-off of NinjaDesc.

\begin{figure}[!t]
    \vspace{-0pt}
    \centering
    \begin{subfigure}[t]{\columnwidth}
        \centering
        \includegraphics[width=\linewidth, trim={
            8pt, 0pt, 30pt, 6pt}, clip]{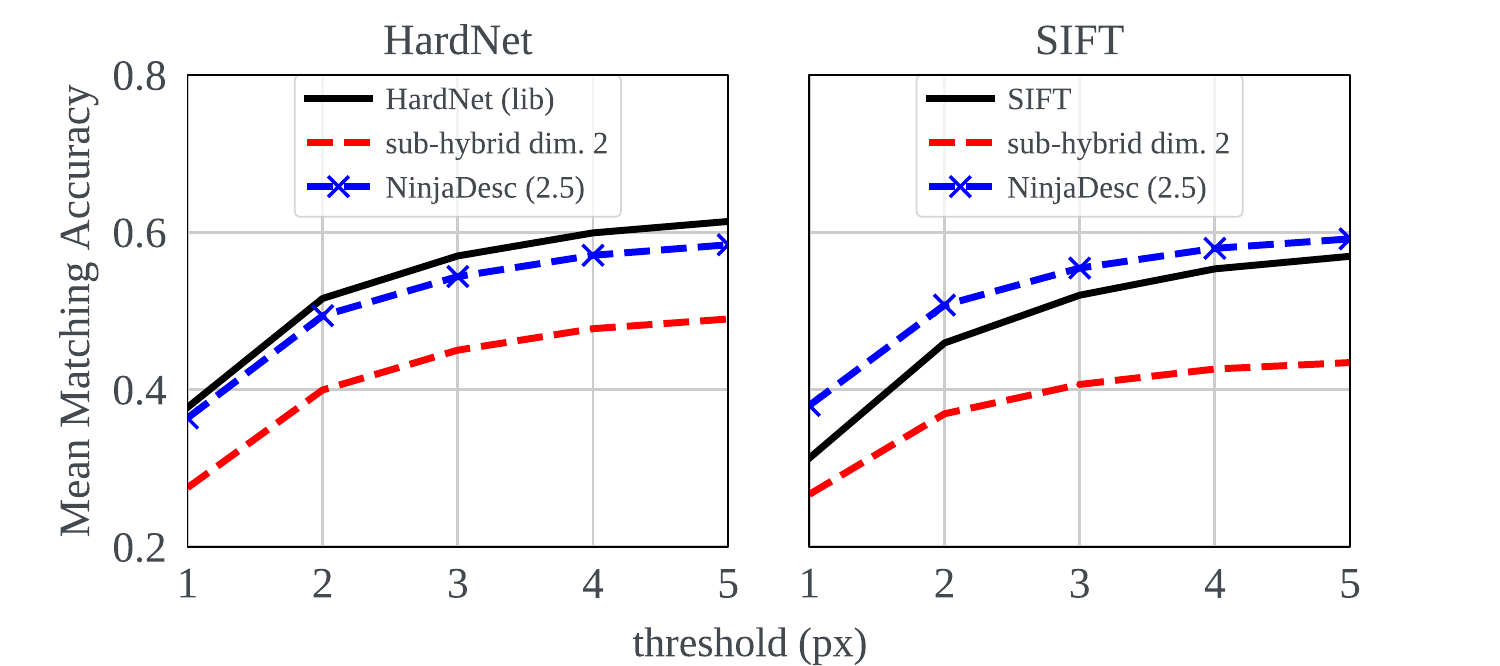}
        \vspace{-5pt}
        \subcaption{\footnotesize{Mean matching accuracy on HPatches~\cite{balntas2017hpatches} sequences. We compare NinjaDesc ($\lambda=2.5$) to sub-hybrid lifting (dim. 2) in Dusmanu \etal~\cite{dusmanu2020privacy}.}}
        \label{subfig:hsequences}
    \end{subfigure}
    \begin{subfigure}[b]{\columnwidth}
        \centering
        \vspace{5pt}
        \includegraphics[width=\linewidth, trim={
            5pt, 0pt, 35pt, 20pt}, clip]{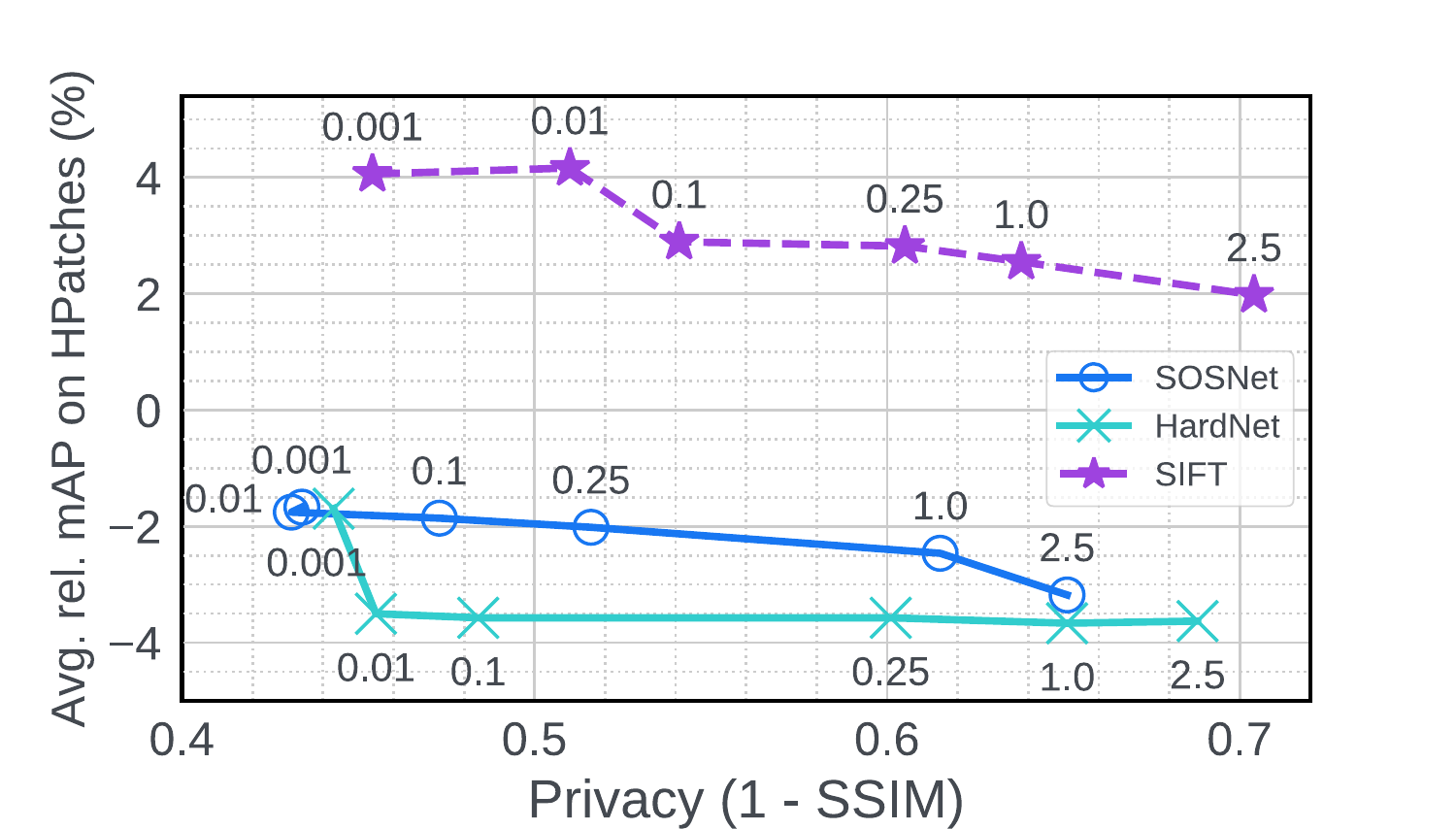}
        \vspace{-10pt}
        \subcaption{\footnotesize{For each descriptor we select NinjaDesc with varying \losslambda values (annotated next to data points), and compare their utility \textit{relative} to the raw descriptor \vs \privacy.}}
        \label{subfig:trade-off}
        \vspace{-7pt}
    \end{subfigure}
    \vspace{-8pt}
    \caption{Utility \vs privacy trade-off analyses.}
    \label{fig:trade-off}
    \vspace{-6pt}
\end{figure}

First, in Fig.~\ref{subfig:hsequences} we evaluate the mean matching accuracy (MMA) of NinjaDesc at the highest \losslambda $\lambda=2.5$, for both HardNet~\cite{mishchuk2017hardnet} and SIFT~\cite{lowe2004sift}, on the HPatches sequences~\cite{balntas2017hpatches} and compare that with the sub-hybrid lifting method by Dusmanu \etal~\cite{dusmanu2020privacy} with low privacy level (dim. 2).
Even at a higher privacy level, NinjaDesc significantly outperforms sub-hybrid lifting for both types of descriptors. 
For NinjaDesc, the drop in MMA \wrt to HardNet is also minimal, and even increases \wrt SIFT. %  (which is elaborated in the next paragraph).

Second, in Fig.~\ref{subfig:trade-off} we perform a detailed utility \vs privacy trade-off analysis on NinjaDesc for all three base descriptors.
The $y$-axis is the average difference in NinjaDesc's mAP across the three tasks in HPatches in Fig.~\ref{fig:hpatches_results}, and the $x$-axis is the privacy measured by 1\,-\,SSIM \cite{dangwal21}.
We plot the results varying the privacy parameter. % $(\lambda = \{0.001, 0.01, 0.1, 0.25, 1.0, 2.5\})$.
For SOSNet and HardNet, the drop in utility ($<4\%$) is a magnitude less than the gain in privacy ($30\%$), indicating an optimal trade-off.
%there is a clear trade-off as we see a drop in utility with increasing privacy.
%However, the trade-off is optimal as the drop in utility is a magnitude smaller than the gain in privacy.
Interestingly, for SIFT we see a net gain in utility for all $\lambda$ (positive values in the $y$-axis). This is due to the SOSNet-like utility training, improving the \textit{verification} and \textit{retrieval} of NinjaDesc beyond the handcrafted SIFT. Full HPatches results for HardNet and SIFT are in the supplementary.

\section{Limitations}
% For all experiments in this paper, we use a fixed number of keypoints (1000) detected for all images. This already incurs very high compute cost in training, therefore we have not explored the possibility if NinjaDesc generalizes to cases where the density of keypoints is much higher.
NinjaDesc only affects the descriptors, and not the keypoint locations. Therefore, it does not prevent inferring scene structures from the patterns of keypoint locations themselves \cite{linecloud,Luo_2019_CVPR}. Also, some level of structure can still be revealed where keypoints are very dense, \eg the venetian blinds in the second example of Fig.~\ref{fig:recon_faces}.
% This method can be complimentary to keypoint obfuscating methods \cite{}.
%Future work could consider concealing keypoint locations as well, while still applicable to downstream tasks that normally requires such information (\eg pose estimation, image warping \etc).
%The adversarial learning process could be filtering out information in the descriptors beyond the needs for matching, which is crucial for reconstruction.
%The spatial correlation might have also been tampered, without affecting matching.

\section{Conclusions}
We introduced a novel adversarial learning framework for visual descriptors to prevent reconstructing original input image content from the descriptors. We experimentally validated that the obtained descriptors deteriorate the descriptor inversion quality with only marginal drop in utility. % using the standard benchmarks for descriptor matching. 
We also empirically demonstrated that we can control the trade-off between utility and non-invertibility using our framework, by changing a single parameter that weighs the adversarial loss. The ablation studies using different types of visual descriptors and image reconstruction network architectures demonstrate the generalizability of our method. 
Our proposed pipeline can enhance the security of computer vision systems that use visual descriptors, and has great potential to be extended for other applications beyond local descriptor encoding. Our observation suggests that the visual descriptors contain more information than what is needed for matching, which is removed by the adversarial learning process. It opens up a new opportunity in general representation learning for obtaining representations with only necessary information to preserve privacy. %for the target task to preserve privacy. 

%Our proposed pipeline can enhance the security of computer vision systems that use visual descriptors, and has potential to be extended for other applications beyond local descriptor encoding.Future work may include a deep dive into explainability of the better trade-off brought by the adversarial learning process. We speculate that the adversarial learning process could be filtering out information in the descriptors beyond the needs for matching, or tampering with the spatial correlation of descriptors required for reconstruction.
%

\parheader{Acknowledgement.} This work was supported by the Chist-Era EPSRC IPALM EP/S032398/1 grant.

%%%%%%%%% REFERENCES
{\small
\bibliographystyle{ieee_fullname}
\bibliography{references}

\begin{thebibliography}{10}\itemsep=-1pt

\bibitem{agarwal2011building}
Sameer Agarwal, Yasutaka Furukawa, Noah Snavely, Ian Simon, Brian Curless,
  Steven~M Seitz, and Richard Szeliski.
\newblock Building rome in a day.
\newblock {\em Communications of the ACM}, 2011.

\bibitem{alahi2012freak}
Alexandre Alahi, Raphael Ortiz, and Pierre Vandergheynst.
\newblock Freak: Fast retina keypoint.
\newblock In {\em CVPR}, 2012.

\bibitem{arandjelovic2016netvlad}
Relja Arandjelovi{\'c}, Petr Gronat, Akihiko Torii, Tomas Pajdla, and Josef
  Sivic.
\newblock Net{VLAD}: {CNN} architecture for weakly supervised place
  recognition.
\newblock In {\em CVPR}, 2016.

\bibitem{arandjelovic2014dislocation}
Relja Arandjelovi{\'c} and Andrew Zisserman.
\newblock {DisLocation}: Scalable descriptor distinctiveness for location
  recognition.
\newblock In {\em ACCV}, 2014.

\bibitem{baik2020domain}
Sungyong Baik, Hyo~Jin Kim, Tianwei Shen, Eddy Ilg, Kyoung~Mu Lee, and
  Christopher Sweeney.
\newblock Domain adaptation of learned featuresfor visual localization.
\newblock In {\em BMVC}, 2020.

\bibitem{balntas2017hpatches}
Vassileios Balntas, Karel Lenc, Andrea Vedaldi, and Krystian Mikolajczyk.
\newblock {HPatches}: A benchmark and evaluation of handcrafted and learned
  local descriptors.
\newblock In {\em CVPR}, 2017.

\bibitem{barroso-laguna2019keynet}
Axel Barroso-Laguna, Edgar Riba, Daniel Ponsa, and Krystian Mikolajczyk.
\newblock {Key.Net}: Keypoint detection by handcrafted and learned cnn filters.
\newblock In {\em ICCV}, 2019.

\bibitem{calonder2010brief}
Michael Calonder, Vincent Lepetit, Christoph Strecha, and Pascal Fua.
\newblock {BRIEF}: Binary robust independent elementary features.
\newblock In {\em ECCV}, 2010.

\bibitem{chelani2021howprivacypreserving}
Kunal Chelani, Fredrik Kahl, and Torsten Sattler.
\newblock How privacy-preserving are line clouds? {R}ecovering scene details
  from 3d lines.
\newblock In {\em CVPR}, 2021.

\bibitem{d2013bits}
Emmanuel d'Angelo, Laurent Jacques, Alexandre Alahi, and Pierre Vandergheynst.
\newblock From bits to images: Inversion of local binary descriptors.
\newblock {\em TPAMI}, 36(5):874--887, 2013.

\bibitem{dangwal21}
Deeksha Dangwal, Vincent~T. Lee, Hyo~Jin Kim, Tianwei Shen, Meghan Cowan, Rajvi
  Shah, Caroline Trippel, Brandon Reagen, Timothy Sherwood, Vasileios Balntas,
  Armin Alaghi, and Eddy Ilg.
\newblock Analysis and mitigations of reverse engineering attacks on local
  feature descriptors.
\newblock In {\em BMVC}, 2021.

\bibitem{deng2009ImageNet}
Jia Deng, Wei Dong, Richard Socher, Li-Jia Li, {Kai Li}, and {Li Fei-Fei}.
\newblock {ImageNet}: A large-scale hierarchical image database.
\newblock In {\em CVPR}, 2009.

\bibitem{detone2018superpoint}
Daniel DeTone, Tomasz Malisiewicz, and Andrew Rabinovich.
\newblock {SuperPoint}: Self-supervised interest point detection and
  description.
\newblock In {\em CVPR Workshops}, 2018.

\bibitem{DFDC}
Brian Dolhansky, Joanna Bitton, Ben Pflaum, Jikuo Lu, Russ Howes, Menglin Wang,
  and Cristian Canton{-}Ferrer.
\newblock The deepfake detection challenge dataset.
\newblock {\em CoRR}, abs/2006.07397, 2020.

\bibitem{dong2015distributed}
Jing Dong, Erik Nelson, Vadim Indelman, Nathan Michael, and Frank Dellaert.
\newblock Distributed real-time cooperative localization and mapping using an
  uncertainty-aware expectation maximization approach.
\newblock In {\em ICRA}, 2015.

\bibitem{dosovitskiy2016inverting}
Alexey Dosovitskiy and Thomas Brox.
\newblock Inverting visual representations with convolutional networks.
\newblock In {\em CVPR}, 2016.

\bibitem{dusmanu2019d2net}
Mihai Dusmanu, Ignacio Rocco, Tomas Pajdla, Marc Pollefeys, Josef Sivic,
  Akihiko Torii, and Torsten Sattler.
\newblock {D2-Net}: A trainable cnn for joint detection and description of
  local features.
\newblock In {\em CVPR}, 2019.

\bibitem{dusmanu2020privacy}
Mihai Dusmanu, Johannes~L Sch{\"o}nberger, Sudipta~N Sinha, and Marc Pollefeys.
\newblock Privacy-preserving visual feature descriptors through adversarial
  affine subspace embedding.
\newblock In {\em CVPR}, 2021.

\bibitem{erkin2009privacy}
Zekeriya Erkin, Martin Franz, Jorge Guajardo, Stefan Katzenbeisser, Inald
  Lagendijk, and Tomas Toft.
\newblock Privacy-preserving face recognition.
\newblock In {\em International symposium on privacy enhancing technologies
  symposium}, 2009.

\bibitem{geppert2020privacy}
Marcel Geppert, Viktor Larsson, Pablo Speciale, Johannes~L Sch{\"o}nberger, and
  Marc Pollefeys.
\newblock Privacy preserving structure-from-motion.
\newblock In {\em ECCV}, 2020.

\bibitem{geppert2021privacy}
Marcel Geppert, Viktor Larsson, Pablo Speciale, Johannes~L Schonberger, and
  Marc Pollefeys.
\newblock Privacy preserving localization and mapping from uncalibrated
  cameras.
\newblock In {\em CVPR}, 2021.

\bibitem{goesele2007ubc}
Michael Goesele, Noah Snavely, Brian Curless, Hugues Hoppe, and Steven~M.
  Seitz.
\newblock Multi-view stereo for community photo collections.
\newblock In {\em CVPR}, 2007.

\bibitem{goodfellow2014generative}
Ian~J. Goodfellow, Jean Pouget-Abadie, Mehdi Mirza, Bing Xu, David
  Warde-Farley, Sherjil Ozair, Aaron Courville, and Yoshua Bengio.
\newblock Generative adversarial networks.
\newblock In {\em NIPS}, 2014.

\bibitem{hare2012efficient}
Sam Hare, Amir Saffari, and Philip~HS Torr.
\newblock Efficient online structured output learning for keypoint-based object
  tracking.
\newblock In {\em CVPR}, 2012.

\bibitem{harris_corner}
Christopher~G Harris, Mike Stephens, et~al.
\newblock A combined corner and edge detector.
\newblock In {\em Alvey vision conference}, volume~15, 1988.

\bibitem{he2016ResNet}
Kaiming He, Xiangyu Zhang, Shaoqing Ren, and Jian Sun.
\newblock Deep residual learning for image recognition.
\newblock In {\em CVPR}, 2016.

\bibitem{hinojosa2021learning}
Carlos Hinojosa, Juan~Carlos Niebles, and Henry Arguello.
\newblock Learning privacy-preserving optics for human pose estimation.
\newblock In {\em ICCV}, 2021.

\bibitem{kapture2020}
Martin Humenberger, Yohann Cabon, Nicolas Guerin, Julien Morat, Jérôme
  Revaud, Philippe Rerole, Noé Pion, Cesar de Souza, Vincent Leroy, and
  Gabriela Csurka.
\newblock Robust image retrieval-based visual localization using kapture, 2020.

\bibitem{jegou2008hamming}
Herv{\'e} J{\'e}gou, Matthijs Douze, and Cordelia Schmid.
\newblock Hamming embedding and weak geometry consistency for large scale image
  search.
\newblock In {\em ECCV}, 2008.

\bibitem{jin2021image}
Yuhe Jin, Dmytro Mishkin, Anastasiia Mishchuk, Jiri Matas, Pascal Fua,
  Kwang~Moo Yi, and Eduard Trulls.
\newblock Image matching across wide baselines: From paper to practice.
\newblock {\em IJCV}, 2021.

\bibitem{kato2014image}
Hiroharu Kato and Tatsuya Harada.
\newblock Image reconstruction from bag-of-visual-words.
\newblock In {\em CVPR}, 2014.

\bibitem{kim2017learned}
Hyo~Jin Kim, Enrique Dunn, and Jan-Michael Frahm.
\newblock Learned contextual feature reweighting for image geo-localization.
\newblock In {\em CVPR}, 2017.

\bibitem{adam_optimizer}
Diederik~P Kingma and Jimmy Ba.
\newblock Adam: A method for stochastic optimization.
\newblock In {\em ICLR}, 2015.

\bibitem{krizhevsky2017imagenet}
Alex Krizhevsky, Ilya Sutskever, and Geoffrey~E Hinton.
\newblock {ImageNet} classification with deep convolutional neural networks.
\newblock {\em Communications of the ACM}, 2017.

\bibitem{li2018megadepth}
Zhengqi Li and Noah Snavely.
\newblock {MegaDepth}: Learning single-view depth prediction from internet
  photos.
\newblock In {\em CVPR}, 2018.

\bibitem{liu2010sift}
Ce Liu, Jenny Yuen, and Antonio Torralba.
\newblock {SIFT} flow: Dense correspondence across scenes and its applications.
\newblock {\em TPAMI}, 2010.

\bibitem{lowe2004sift}
David~G. Lowe.
\newblock Distinctive image features from scale-invariant keypoints.
\newblock In {\em IJCV}, 2004.

\bibitem{Luo_2019_CVPR}
Zixin Luo, Tianwei Shen, Lei Zhou, Jiahui Zhang, Yao Yao, Shiwei Li, Tian Fang,
  and Long Quan.
\newblock {ContextDesc}: Local descriptor augmentation with cross-modality
  context.
\newblock In {\em CVPR}, 2019.

\bibitem{mahendran2015understanding}
Aravindh Mahendran and Andrea Vedaldi.
\newblock Understanding deep image representations by inverting them.
\newblock In {\em CVPR}, 2015.

\bibitem{mei2011rslam}
Christopher Mei, Gabe Sibley, Mark Cummins, Paul Newman, and Ian Reid.
\newblock Rslam: A system for large-scale mapping in constant-time using
  stereo.
\newblock {\em IJCV}, 2011.

\bibitem{mishchuk2017hardnet}
Anastasiya Mishchuk, Dmytro Mishkin, Filip Radenovi{\'c}, and Ji{\u{r}}i Matas.
\newblock Working hard to know your neighbor's margins: Local descriptor
  learning loss.
\newblock In {\em NIPS}, 2017.

\bibitem{mur-artal2016orb-slam}
Raul Mur-Artal, J.~M.~M. Montiel, and Juan~D. Tardos.
\newblock {ORB-SLAM}: A versatile and accurate monocular slam system.
\newblock {\em IEEE Transactions on Robotics}, 31(5):1147–1163, Oct 2015.

\bibitem{mur2017orb}
Raul Mur-Artal and Juan~D Tard{\'o}s.
\newblock {ORB-SLAM}2: An open-source slam system for monocular, stereo, and
  {RGB-D} cameras.
\newblock {\em IEEE Transactions on Robotics}, 2017.

\bibitem{nebehay2014consensus}
Georg Nebehay and Roman Pflugfelder.
\newblock Consensus-based matching and tracking of keypoints for object
  tracking.
\newblock In {\em WACV}, 2014.

\bibitem{newcombe2011dtam}
Richard~A. Newcombe, Steven~J. Lovegrove, and Andrew~J. Davison.
\newblock {DTAM}: Dense tracking and mapping in real-time.
\newblock In {\em ICCV}, 2011.

\bibitem{ng2020solar}
Tony Ng, Vassileios Balntas, Yurun Tian, and Krystian Mikolajczyk.
\newblock {SOLAR}: Second-order loss and attention for image retrieval.
\newblock In {\em ECCV}, 2020.

\bibitem{noh2017delf}
Hyeonwoo Noh, Andr{\'e} Araujo, Jack Sim, Tobias Weyand, and Bohyung Han.
\newblock Image retrieval with deep local features and attention-based
  keypoints.
\newblock In {\em ICCV}, 2017.

\bibitem{ojala2002multiresolution}
Timo Ojala, Matti Pietikainen, and Topi Maenpaa.
\newblock Multiresolution gray-scale and rotation invariant texture
  classification with local binary patterns.
\newblock {\em TPAMI}, 2002.

\bibitem{Oth_2013_CVPR}
Luc Oth, Paul Furgale, Laurent Kneip, and Roland Siegwart.
\newblock Rolling shutter camera calibration.
\newblock In {\em CVPR}, 2013.

\bibitem{paszke2019PyTorch}
Adam Paszke, Sam Gross, Francisco Massa, Adam Lerer, James Bradbury, Gregory
  Chanan, Trevor Killeen, Zeming Lin, Natalia Gimelshein, Luca Antiga, Alban
  Desmaison, Andreas Kopf, Edward Yang, Zachary DeVito, Martin Raison, Alykhan
  Tejani, Sasank Chilamkurthy, Benoit Steiner, Lu Fang, Junjie Bai, and Soumith
  Chintala.
\newblock {PyTorch: A}n imperative style, high-performance deep learning
  library.
\newblock In {\em NeurIPS}, 2019.

\bibitem{pernici2013object}
Federico Pernici and Alberto Del~Bimbo.
\newblock Object tracking by oversampling local features.
\newblock {\em TPAMI}, 2013.

\bibitem{pittaluga2019learning}
Francesco Pittaluga, Sanjeev Koppal, and Ayan Chakrabarti.
\newblock Learning privacy preserving encodings through adversarial training.
\newblock In {\em WACV}, 2019.

\bibitem{invsfm}
Francesco Pittaluga, Sanjeev~J Koppal, Sing~Bing Kang, and Sudipta~N Sinha.
\newblock Revealing scenes by inverting structure from motion reconstructions.
\newblock In {\em CVPR}, 2019.

\bibitem{porav2018adversarial}
Horia Porav, Will Maddern, and Paul Newman.
\newblock Adversarial training for adverse conditions: Robust metric
  localisation using appearance transfer.
\newblock In {\em ICRA}, 2018.

\bibitem{revaud2019aploss}
Jerome Revaud, Jon Almaz{\'{a}}n, Rafael Sampaio~de Rezende, and C{\'e}sar
  Roberto~de Souza.
\newblock Learning with average precision: Training image retrieval with a
  listwise loss.
\newblock In {\em ICCV}, 2019.

\bibitem{revaud2019r2d2}
Jerome Revaud, Philippe Weinzaepfel, César De~Souza, Noe Pion, Gabriela
  Csurka, Yohann Cabon, and Martin Humenberger.
\newblock {R2D2}: Repeatable and reliable detector and descriptor.
\newblock In {\em NeurIPS}, 2019.

\bibitem{riba2020kornia}
Edgar Riba, Dmytro Mishkin, Daniel Ponsa, Ethan Rublee, and Gary Bradski.
\newblock Kornia: an open source differentiable computer vision library for
  {PyTorch}.
\newblock In {\em WACV}, 2020.

\bibitem{unet}
Olaf Ronneberger, Philipp Fischer, and Thomas Brox.
\newblock {U-Net}: Convolutional networks for biomedical image segmentation.
\newblock In {\em MICCAI}. Springer, 2015.

\bibitem{roy2019mitigating}
Proteek~Chandan Roy and Vishnu~Naresh Boddeti.
\newblock Mitigating information leakage in image representations: A maximum
  entropy approach.
\newblock In {\em CVPR}, 2019.

\bibitem{sadeghi2009efficient}
Ahmad-Reza Sadeghi, Thomas Schneider, and Immo Wehrenberg.
\newblock Efficient privacy-preserving face recognition.
\newblock In {\em International Conference on Information Security and
  Cryptology}, 2009.

\bibitem{sarlin20superglue}
Paul-Edouard Sarlin, Daniel DeTone, Tomasz Malisiewicz, and Andrew Rabinovich.
\newblock {SuperGlue}: Learning feature matching with graph neural networks.
\newblock In {\em CVPR}, 2020.

\bibitem{sattler2017active}
Torsten Sattler, Bastian Leibe, and Leif Kobbelt.
\newblock Efficient \& effective prioritized matching for large-scale
  image-based localization.
\newblock {\em TPAMI}, 39(9):1744--1756, 2017.

\bibitem{sattler2018benchmarking}
Torsten Sattler, Will Maddern, Carl Toft, Akihiko Torii, Lars Hammarstrand,
  Erik Stenborg, Daniel Safari, Masatoshi Okutomi, Marc Pollefeys, Josef Sivic,
  Fredrik Kahl, and Tomas Pajdla.
\newblock Benchmarking 6dof outdoor visual localization in changing conditions.
\newblock In {\em CVPR}, 2018.

\bibitem{schonberger2017comparative}
Johannes~L Schonberger, Hans Hardmeier, Torsten Sattler, and Marc Pollefeys.
\newblock Comparative evaluation of hand-crafted and learned local features.
\newblock In {\em CVPR}, 2017.

\bibitem{schonberger2016colmap}
Johannes~L. Schönberger and Jan-Michael Frahm.
\newblock Structure-from-motion revisited.
\newblock In {\em CVPR}, 2016.

\bibitem{shibuya2020privacy}
Mikiya Shibuya, Shinya Sumikura, and Ken Sakurada.
\newblock Privacy preserving visual {SLAM}.
\newblock In {\em ECCV}, 2020.

\bibitem{simeoni2019local}
Oriane Sim{\'e}oni, Yannis Avrithis, and Ondrej Chum.
\newblock Local features and visual words emerge in activations.
\newblock In {\em CVPR}, 2019.

\bibitem{simonyan2015VGG}
Karen Simonyan and Andrew Zisserman.
\newblock Very deep convolutional networks for large-scale image recognition.
\newblock In {\em ICLR}, 2015.

\bibitem{sivic2003video}
Josef Sivic and Andrew Zisserman.
\newblock Video google: A text retrieval approach to object matching in videos.
\newblock In {\em ICCV}, 2003.

\bibitem{linecloud}
Pablo Speciale, Johannes~L Schonberger, Sing~Bing Kang, Sudipta~N Sinha, and
  Marc Pollefeys.
\newblock Privacy preserving image-based localization.
\newblock In {\em CVPR}, 2019.

\bibitem{speciale2019privacy}
Pablo Speciale, Johannes~L Schonberger, Sudipta~N Sinha, and Marc Pollefeys.
\newblock Privacy preserving image queries for camera localization.
\newblock In {\em CVPR}, 2019.

\bibitem{sweeney2015theia}
Chris Sweeney, Tobias Hollerer, and Matthew Turk.
\newblock Theia: A fast and scalable structure-from-motion library.
\newblock In {\em Proceedings of the 23rd ACM International Conference on
  Multimedia}, MM '15, page 693–696, 2015.

\bibitem{tian2020hynet}
Yurun Tian, Axel Barroso-Laguna, Tony Ng, Vassileios Balntas, and Krystian
  Mikolajczyk.
\newblock {HyNet}: Learning local descriptor with hybrid similarity measure and
  triplet loss.
\newblock In {\em NeurIPS}, 2020.

\bibitem{tian2017l2net}
Yurun Tian, Bin Fan, and Fuchao Wu.
\newblock {L2-Net}: Deep learning of discriminative patch descriptor in
  {E}uclidean space.
\newblock In {\em CVPR}, 2017.

\bibitem{tian2019sosnet}
Yurun Tian, Xin Yu, Bin Fan, Wu. Fuchao, Huub Heijnen, and Vassileios Balntas.
\newblock {SOSNet}: Second order similarity regularization for local descriptor
  learning.
\newblock In {\em CVPR}, 2019.

\bibitem{toft2020long}
Carl Toft, Will Maddern, Akihiko Torii, Lars Hammarstrand, Erik Stenborg,
  Daniel Safari, Masatoshi Okutomi, Marc Pollefeys, Josef Sivic, Tomas Pajdla,
  Fredrik Kahl, and Torsten Sattler.
\newblock Long-term visual localization revisited.
\newblock {\em TPAMI}, 2020.

\bibitem{toft2020single}
Carl Toft, Daniyar Turmukhambetov, Torsten Sattler, Fredrik Kahl, and Gabriel~J
  Brostow.
\newblock Single-image depth prediction makes feature matching easier.
\newblock In {\em ECCV}, 2020.

\bibitem{tolias2013smk}
Giorgos Tolias, Yannis Avrithis, and Herv{\'e} J{\'e}gou.
\newblock To aggregate or not to aggregate: Selective match kernels for image
  search.
\newblock In {\em ICCV}, 2013.

\bibitem{tolias2020learning}
Giorgos Tolias, Tomas Jenicek, and Ond{\v{r}}ej Chum.
\newblock Learning and aggregating deep local descriptors for instance-level
  recognition.
\newblock In {\em ECCV}, 2020.

\bibitem{hoggles}
Carl Vondrick, Aditya Khosla, Tomasz Malisiewicz, and Antonio Torralba.
\newblock Hoggles: Visualizing object detection features.
\newblock In {\em ICCV}, 2013.

\bibitem{weinzaepfel2011reconstructing}
Philippe Weinzaepfel, Herv{\'e} J{\'e}gou, and Patrick P{\'e}rez.
\newblock Reconstructing an image from its local descriptors.
\newblock In {\em CVPR}, 2011.

\bibitem{xiao2020adversarial}
Taihong Xiao, Yi-Hsuan Tsai, Kihyuk Sohn, Manmohan Chandraker, and Ming-Hsuan
  Yang.
\newblock Adversarial learning of privacy-preserving and task-oriented
  representations.
\newblock In {\em AAAI}, 2020.

\bibitem{xie2017controllable}
Qizhe Xie, Zihang Dai, Yulun Du, Eduard Hovy, and Graham Neubig.
\newblock Controllable invariance through adversarial feature learning.
\newblock In {\em NIPS}, 2017.

\bibitem{yonetani2017privacy}
Ryo Yonetani, Vishnu Naresh~Boddeti, Kris~M Kitani, and Yoichi Sato.
\newblock Privacy-preserving visual learning using doubly permuted homomorphic
  encryption.
\newblock In {\em ICCV}, 2017.

\bibitem{zhang2020aachenv_1_1}
Zichao Zhang, Torsten Sattler, and Davide Scaramuzza.
\newblock Reference pose generation for long-term visual localization via
  learned features and view synthesis.
\newblock {\em IJCV}, 2020.

\bibitem{zhu2006fast}
Qiang Zhu, Mei-Chen Yeh, Kwang-Ting Cheng, and Shai Avidan.
\newblock Fast human detection using a cascade of histograms of oriented
  gradients.
\newblock In {\em CVPR}, 2006.

\end{thebibliography}
}

\clearpage
\section*{Supplementary material}
\appendix
We first provide a comparison of our NinjaDesc and the base descriptor on the 3D reconstruction task using SfM (Sec.~\ref{supp-sec:sfm}). Next, we report the full HPatches results using HardNet~\cite{mishchuk2017hardnet} and SIFT~\cite{lowe2004sift} as the base descriptors (Sec.~\ref{supp-sec:hpatches}). In addition to our results on Aachen-Day-Night v1.1 in the main paper, we also provide our results on Aachen-Day-Night v1.0 (Sec.~\ref{supp-sec:aachen_v1_0}). Finally, we illustrate the detailed architecture for the inverse models (Sec.~\ref{supp-sec:arch}).

\section{3D Reconstruction}
\label{supp-sec:sfm}
Table \ref{tab:lfe} shows a quantitative comparison of our content-concealing NinjaDesc and the base descriptor SOSNet~\cite{tian2019sosnet} on the SfM reconstruction task using the landmarks dataset for local feature benchmarking~\cite{schonberger2017comparative}. As can be seen, decrease in the performance for our content-concealing NinjaDesc is only marginal for all metrics. % show competitive performance
\begin{table}[h!]
\resizebox{.48\textwidth}{!}{
\setlength{\tabcolsep}{2.0pt}
\renewcommand{\arraystretch}{1.5}
	\scriptsize
		\centering
		\begin{tabular}{c c c c c c c}
			\toprule
			\multirow{2}{*}{Dataset} & \multirow{2}{*}{Method} & \multirowcell{2}{Reg.\\images} & \multirowcell{2}{Sparse\\points} &
			\multirowcell{2}{Obser-\\vations} &
			\multirowcell{2}{Track\\length} & \multirowcell{2}{Reproj.\\error} \\
			& & & & & \\ \midrule
			\multirowcell{3}{\textit{South-} \\ \textit{Building} \\ $128$ images} & SOSNet & 128 & 101,568 & 638,731  & 6.29 & 0.56 \\
			\cmidrule{2-7}
			& NinjaDesc (1.0) & 128 & 105,780 & 652,869 & 6.17 & 0.56 \\
			& NinjaDesc (2.5) & 128 & 105,961 & 653,449 & 6.17 & 0.56 \\ 
			\midrule
			\multirowcell{3}{\textit{Madrid} \\ \textit{Metropolis} \\ $1344$ images} & SOSNet & 572 & 95,733 & 672,836  & 7.03 & 0.62 \\
			\cmidrule{2-7}
			& NinjaDesc (1.0) & 566 & 94,374 & 668,148 & 7.08 & 0.64 \\
			& NinjaDesc (2.5) & 564 & 94,104 & 667,387 & 7.09 & 0.63 \\ 
			\midrule 
			\multirowcell{3}{\textit{Gendarmen-}\\
			\textit{markt} \\ $1463$ images} & SOSNet & 1076  & 246,503  & 1,660,694  & 6.74 & 0.74 \\
			\cmidrule{2-7}
			& NinjaDesc (1.0) & 1087 & 312,469 & 1,901,060 & 6.08 & 0.75 \\
			& NinjaDesc (2.5) & 1030 & 340,144 & 1,871,726 & 5.50 & 0.77 \\ 
			\midrule 
			\multirowcell{3}{\textit{Tower of}\\
			\textit{London} \\ $1463$ images} & SOSNet & 825  & 200,447  & 1,733,994  & 8.65 & 0.62 \\
			\cmidrule{2-7}
			& NinjaDesc (1.0) & 797 & 198,767 & 1,727,785 & 8.69 & 0.62 \\
			& NinjaDesc (2.5) & 837 & 218,888 & 1,792,908 & 8.19 & 0.64 \\ 

%%%%			& HardNet & 459 &  & 7.25 & 0.89 \\
%%%%			& [P] HardNet - dim. 2 & 367 &  & 6.49 & 0.68 \\
%%%%			& [P] HardNet - dim. 4 & 268 & & 6.32 & 0.58 \\ \midrule

%%%%			\multirowcell{6.5}{\textit{Gendarmen-} \\ \textit{markt} \\ $1463$ images} & SIFT & 896 &  & 6.37 & 0.84 \\
%%%%			& [P] SIFT - dim. 2 & 783 & & 5.44 & 0.71 \\
%%%%			& [P] SIFT - dim. 4 & 458 &  & 5.23 &  0.60 \\ \cmidrule{2-6}
%%%%			& HardNet & 999 & & 6.68 & 0.96 \\
%%%%			& [P] HardNet - dim. 2 & 864 & & 5.98 & 0.80 \\
%%%%			& [P] HardNet - dim. 4 & 751 & & 5.50 & 0.69 \\ \midrule
%%%%			\multirowcell{6.5}{\textit{Tower of} \\ \textit{London} \\ $1576$ images} & SIFT & 635 & & 7.78 & 0.70 \\
%%%%			& [P] SIFT - dim. 2 & 525 & & 6.58 & 0.61 \\
%%%%			& [P] SIFT - dim. 4 & 439 & & 6.10 & 0.56 \\ \cmidrule{2-6}
%%%%			& HardNet & 749 & & 7.85 & 0.81 \\
%%%%			& [P] HardNet - dim. 2 & 557 & & 7.19 & 0.68 \\
%%%%			& [P] HardNet - dim. 4 & 498 & & 6.69 & 0.61 \\
			\bottomrule
		\end{tabular}
	}
	\caption{3D reconstruction statistics on the local feature evaluation benchmark~\cite{schonberger2017comparative}. Number in parenthesis is the privacy parameter $\lambda$.}
	\label{tab:lfe}
	\vspace{-0pt}
\end{table}

\section{Full HPatches results for HardNet and SIFT}
\label{supp-sec:hpatches}
\begin{figure*}[!t]
    \vspace{0pt}
    \centering
    \begin{subfigure}[t]{\linewidth}
        \centering
        \includegraphics[width=\linewidth, trim={
            8pt, 0pt, 30pt, 0pt}, clip]{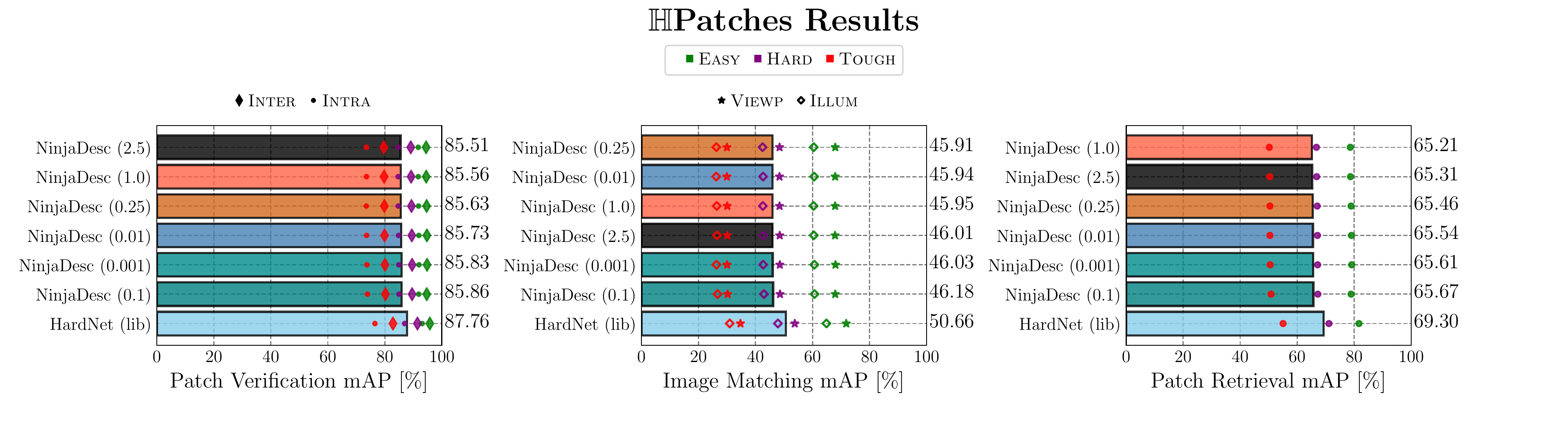}
        \vspace{-20pt}
        \caption{HardNet Base Descriptor}
        \label{supp-subfig:hpatches-hardnet}
    \end{subfigure}
    \begin{subfigure}[b]{\linewidth}
        \centering
        \vspace{8pt}
        \includegraphics[width=\linewidth, trim={
            5pt, 0pt, 35pt, 80pt}, clip]{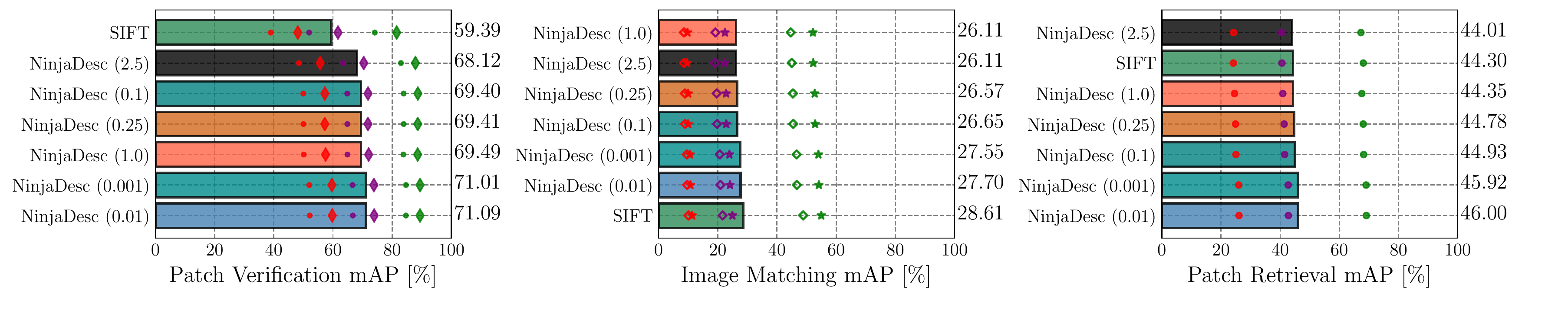}
        \vspace{-20pt}
        \caption{SIFT Base Descriptor}
        \label{supp-subfig:hpatches-sift}
        \vspace{-7pt}
    \end{subfigure}
    \vspace{-0pt}
    \caption{HPatches evaluation results. For each base descriptor (HardNet~\cite{mishchuk2017hardnet} and SIFT~\cite{lowe2004sift}), we compare with NinjaDesc, with 5 different levels of \losslambda $\lambda$ (indicated by the number in parenthesis). All results are from models trained on the \textit{liberty} subset of the UBC patches~\cite{goesele2007ubc} dataset, apart from SIFT which is handcrafted, and we use the Kornia~\cite{riba2020kornia} GPU implementation evaluated on $32\times32$ patches.}
    \label{fig-supp:hpatches}
    \vspace{-0pt}
\end{figure*}

Figure \ref{fig-supp:hpatches} illustrates our full evaluation results on HPatches using HardNet~\cite{mishchuk2017hardnet} and SIFT~\cite{lowe2004sift} as the base descriptors for NinjaDesc, in addition to the results using SOSNet~\cite{tian2019sosnet} provided in the main paper. Similar to the results for SOSNet~\cite{tian2019sosnet}, we observe little drop in accuracy for NinjaDesc overall compared to the original base descriptors, ranging from low ($\lambda=0.1$) to high ($\lambda=2.5$) privacy parameters.

\section{Evaluation on Aachen-Day-Night v1.0}
\label{supp-sec:aachen_v1_0}
In Table 2 of the main paper, we report the result of NinjaDesc on Aachen-Day-Night v1.1 dataset. The v1.1 is updated with more accurate ground-truths compared to the older v1.0. Because Dusmanu~\etal \cite{dusmanu2020privacy} performed evaluation on the v1.0, we also provide our results on v1.0 in Table~\ref{supp-tab:aachen_v1_0} for better comparison.
\begin{table}[h!]
\setlength{\tabcolsep}{2.0pt}
\renewcommand{\arraystretch}{1.1}
	% \footnotesize
	\scriptsize
	\centering
	\resizebox{.48\textwidth}{!}{
	    \begin{tabular}{c c c c c c}
		\toprule
		& & \multirow{3}{*}{Method} & \multicolumn{3}{c}{Accuracy @ Thresholds (\%)} \\
		\multirow{2}{*}{Query} & \multirow{2}{*}{NNs} & & $0.25$m, $2^{\circ}$ & $0.5$m, $5^{\circ}$ & $5.0$m, $10^{\circ}$ \\ 
		\cmidrule{4-6}
		& & \tiny{Base Desc} &  \tiny{~SOS~/~Hard~/~SIFT~}  & \tiny{~SOS~/~Hard~/~SIFT~} & \tiny{~SOS~/~Hard~/~SIFT~}\\
		\midrule
		\multirowcell{8}{Day\\($824$)} & \multirowcell{4.5}{20} 
    	& Raw               & ~85.1\,/\,85.4\,/\,84.3~ & ~92.7\,/\,93.1\,/\,92.7~ & ~97.3\,/\,98.2\,/\,97.6~ \\
		& & $\lambda=0.1$   & ~85.4\,/\,84.7\,/\,82.0~ & ~92.5\,/\,91.9\,/\,91.1~ & ~97.5\,/\,96.8\,/\,96.4~ \\
		& & $\lambda=1.0$   & ~84.7\,/\,84.3\,/\,82.9~ & ~92.4\,/\,91.9\,/\,91.0~ & ~97.2\,/\,96.7\,/\,96.1~ \\
		& & $\lambda=2.5$   & ~84.6\,/\,83.7\,/\,82.5~ & ~92.4\,/\,92.0\,/\,91.0~ & ~97.1\,/\,96.8\,/\,96.0~ \\
		\cmidrule{3-6}
		& \multirowcell{4.5}{50} 
		& Raw               & ~85.9\,/\,86.8\,/\,86.0~ & ~92.5\,/\,93.7\,/\,94.1~ & ~97.3\,/\,98.1\,/\,98.2~ \\
		& & $\lambda=0.1$   & ~85.2\,/\,85.2\,/\,84.2~ & ~92.2\,/\,92.4\,/\,91.4~ & ~97.1\,/\,97.1\,/\,96.6~ \\
		& & $\lambda=1.0$   & ~84.7\,/\,85.7\,/\,83.4~ & ~92.2\,/\,92.6\,/\,91.6~ & ~97.2\,/\,96.7\,/\,96.7~ \\
		& & $\lambda=2.5$   & ~85.6\,/\,85.3\,/\,83.6~ & ~92.7\,/\,91.7\,/\,91.1~ & ~97.3\,/\,96.8\,/\,96.2~ \\
		\cmidrule{2-6}
		\multirowcell{5.5}{Night\\($98$)} & \multirowcell{4.5}{20} & 
		Raw               & ~51.0\,/\,57.2\,/\,55.1~ & ~65.3\,/\,68.4\,/\,67.3~ & ~70.4\,/\,76.5\,/\,74.5~ \\
		& & $\lambda=0.1$ & ~51.0\,/\,45.9\,/\,45.9~ & ~62.2\,/\,56.1\,/\,54.1~ & ~68.4\,/\,62.2\,/\,63.3~ \\
		& & $\lambda=1.0$ & ~50.0\,/\,43.9\,/\,44.9~ & ~62.2\,/\,54.1\,/\,56.1~ & ~66.3\,/\,62.2\,/\,64.3~ \\
		& & $\lambda=2.5$ & ~48.0\,/\,44.9\,/\,44.9~ & ~58.2\,/\,59.2\,/\,52.0~ & ~65.3\,/\,65.3\,/\,62.2~ \\
		\cmidrule{3-6}
		  & \multirowcell{4.5}{50} & 
		Raw                 & ~48.0\,/\,51.0\,/\,54.1~ & ~59.2\,/\,64.3\,/\,65.3~ & ~65.3\,/\,68.4\,/\,74.5~ \\
		& & $\lambda=0.1$   & ~41.8\,/\,39.8\,/\,41.8~ & ~52.0\,/\,51.0\,/\,52.0~ & ~60.2\,/\,56.1\,/\,60.2~ \\
		& &  $\lambda=1.0$  & ~43.9\,/\,39.8\,/\,43.9~ & ~54.1\,/\,50.0\,/\,54.1~ & ~63.3\,/\,58.2\,/\,63.3~ \\
		& & $\lambda=2.5$   & ~42.9\,/\,40.8\,/\,42.9~ & ~52.0\,/\,50.0\,/\,52.0~ & ~61.2\,/\,56.1\,/\,58.2~ \\
		
		\bottomrule
	\end{tabular}
	}
	\vspace{-0pt}
	\caption{Visual localization results on Aachen-Day-Night v1.0~\cite{sattler2018benchmarking}. `Raw' corresponds to the base descriptor in each column, followed by three $\lambda$ vales (0.1, 1.0, 2.5) for NinjaDesc.}
	\label{supp-tab:aachen_v1_0}
	\vspace{-0pt}
\end{table}

\section{Additional content-concealment experiments} 

\begin{figure}[!ht]
    \vspace{-3pt}
    \centering 
    \includegraphics[width=\linewidth, trim={
        10pt, 10pt, 10pt, 10pt}]{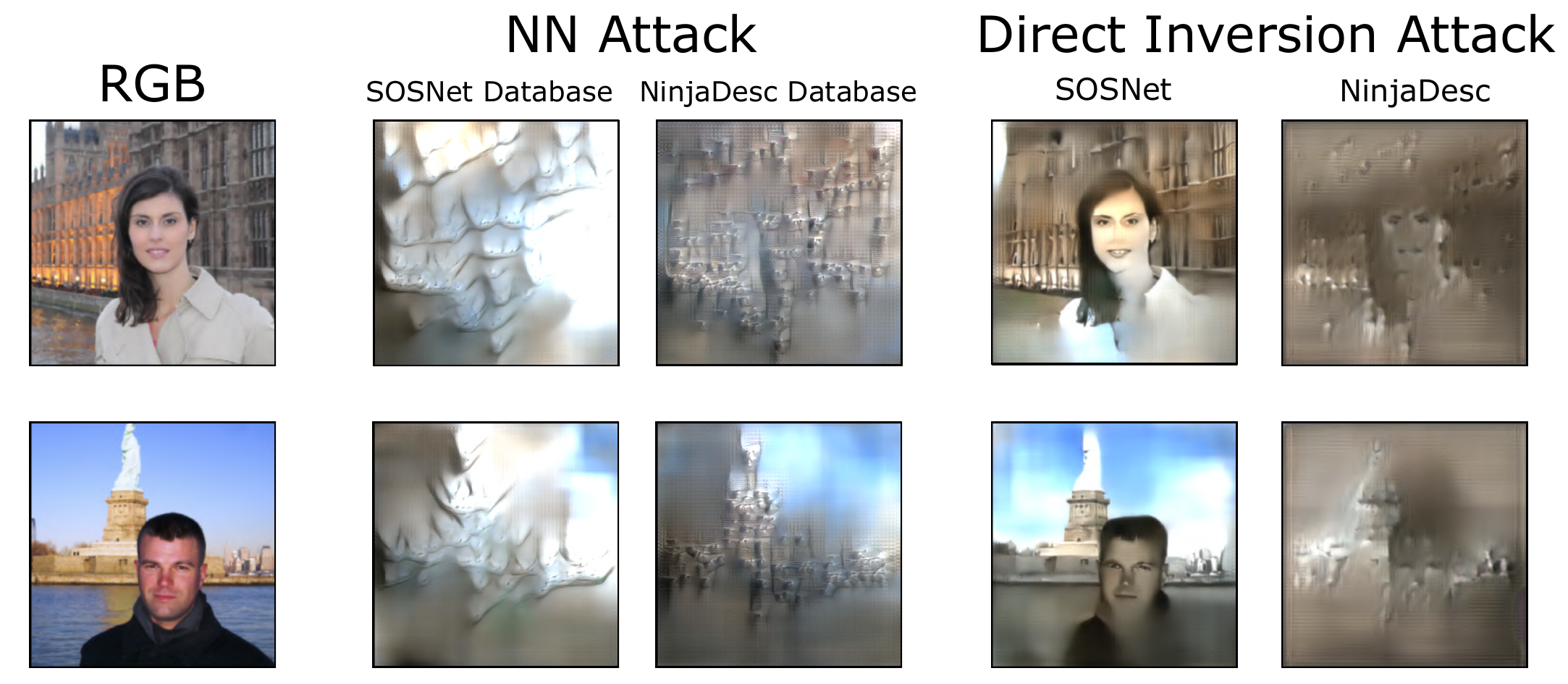}
    \vspace{-17pt}
    \caption{Examples of NN attack. For NN attack, we show results using SOSNet and our NinjaDesc descriptors to form the database.}
    \label{fig:nn_attack}
    \vspace{-13pt}
\end{figure}

\parheader{1. Nearest-neighbour attack.}
Two examples of nearest-neighbour (NN) attack similar to that in \cite{dusmanu2020privacy} using a database of 128,000 existing descriptors are shown in Fig.~\ref{fig:nn_attack}. In both NN attack scenarios, the reconstruction is significantly deteriorated, as it is non-trivial to compute distances between the two spaces, \cf oracle attack analysis below. Note we use $\lambda=2.5$ for all our experiments.

% The oracle attack in [16] is not directly relevant for our method. Unlike [16] where the subspaces lies within the original feature space, our NinjaNet maps the original feature space to a completely new feature space via non-linear transformations. %Thus, it is hard to get any signals about the manifold by computing the distances between two completely different feature spaces, other than the fact that they are not matchable (

\begin{figure}[!ht]
    \vspace{12pt}
    \centering 
    \includegraphics[width=\linewidth, trim={
        20pt, 20pt, 30pt, 20pt}]{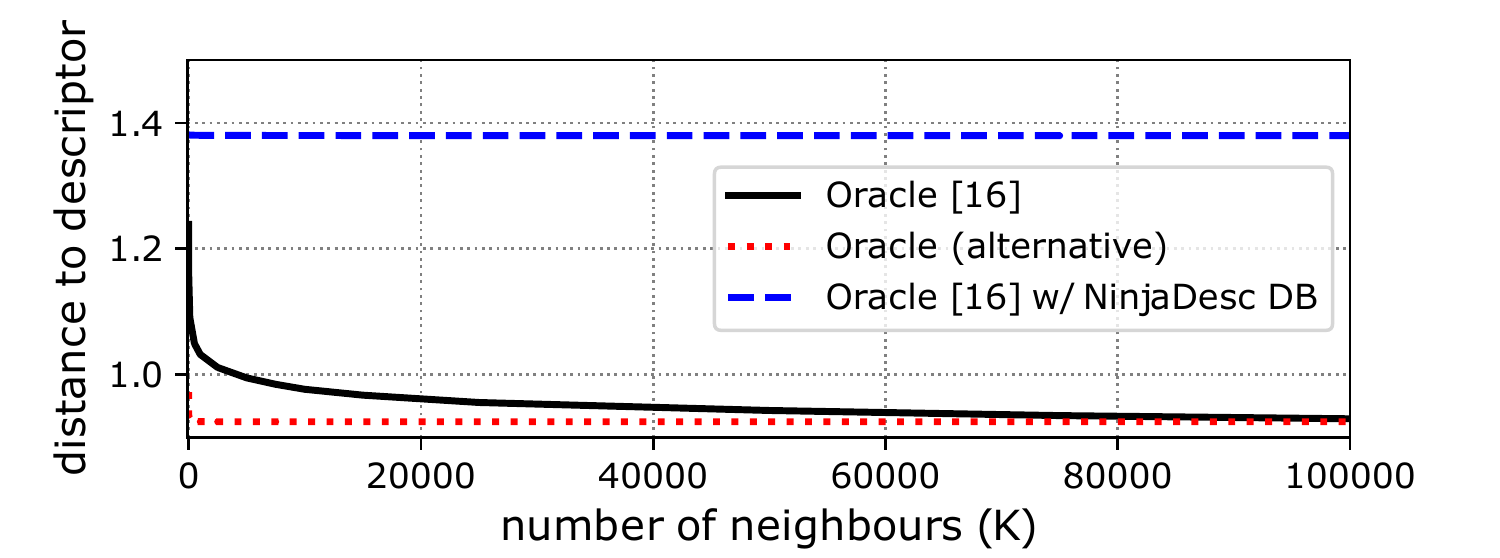}
    \vspace{-8pt}
    \caption{Distances to the original descriptor (SOSNet) of the nearest-neighbour retrieved by three variants of the oracle attack.}
    \label{fig:distances}
    \vspace{-15pt}
\end{figure}

\parheader{2. Oracle attack distance analysis.}
The distances to the original descriptor using the oracle attack following \cite{dusmanu2020privacy} is plotted in black in Fig.~\ref{fig:distances}.
We also show an alternative oracle (\textcolor{red}{red} dotted), which differs from \cite{dusmanu2020privacy} in that the K neighbours are first matched using the NinjaDesc database, then their corresponding SOSNet descriptor pairings are retrieved. 
For completeness, we also plot the results of only using NinjaDesc descriptors as the database (\textcolor{blue}{blue} dashed).

% The oracle attack provides the adversary with an imaginary oracle that, given a set of suspected input descriptors (SOSNet) for a private feature (NinjaDesc), returns the closest one to the original descriptor.
% Following [16], we used the database of 128,000 SOSNet descriptors and their NinjaDesc pairs for retrieving the K closest elements to a query NinjaDesc, and let the oracle to output the closest one to the original SOSNet descriptor. 
%The distances between the original and the SOSNet descriptor varying K is shown in Figure \ref{fig:distances} (SOSNet Database). We show the results for both SOSNet and NinjaDesc databases. 
We observe that the distance decreases as K increases for SOSNet database like Fig. 6 in \cite{dusmanu2020privacy}. 
However, we argue that this alone does not validate manifold folding.
Rather, as K increases we approach the limit of the distance to the real NN of the original (SOSNet) descriptor, regardless of the private (NinjaDesc) representation. % becomes irrelevant.
This limit is achieved by the alternative oracle (\textcolor{red}{red} dotted), where the closest NinjaDesc (\ie the corresponding SOSNet) database descriptor is always retrieved, for most K values. 
If the oracle in \cite{dusmanu2020privacy} uses the NinjaDesc database (\textcolor{blue}{blue} dashed), the distance remains large. This is because unlike \cite{dusmanu2020privacy}, NinjaNet maps the original feature space to a completely new one via learned non-linear transformations, and is thus robust to distance calculation across the two descriptor spaces.

\begin{figure}[!ht]
    \vspace{0pt}
    \centering 
    \includegraphics[width=\linewidth, trim={
        10pt, 20pt, 10pt, 20pt}]{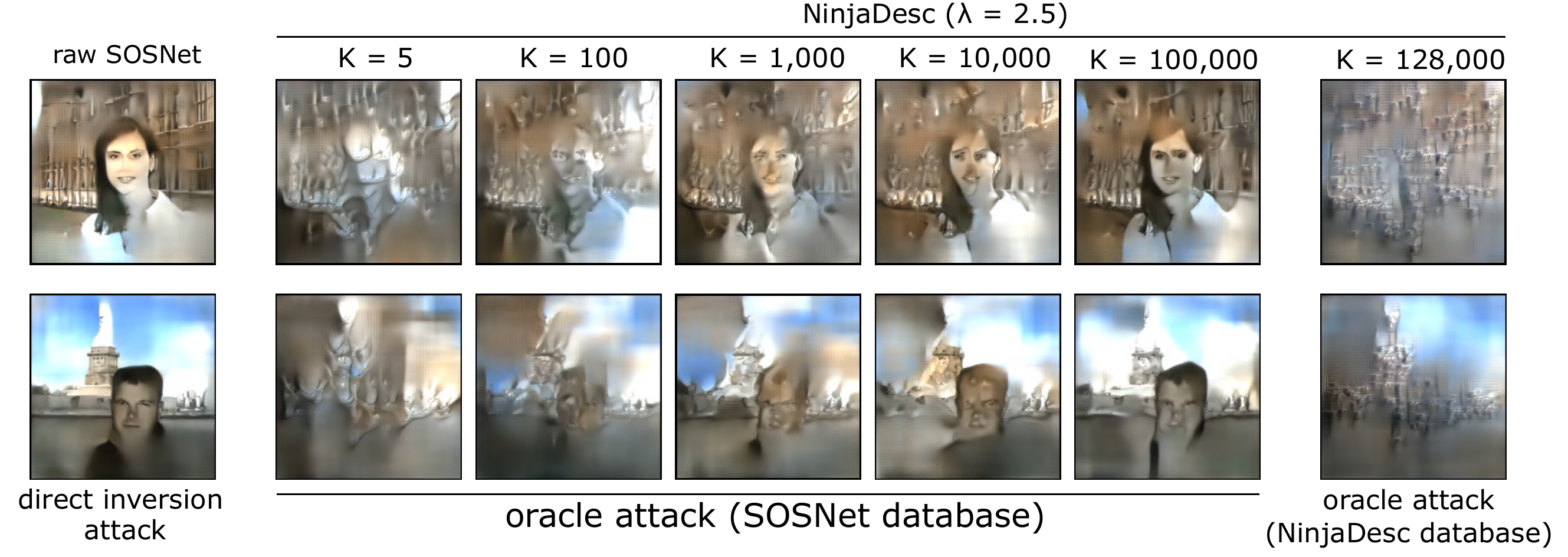}
    \vspace{-15pt}
    \caption{Examples of oracle attack \wrt num. of neighbours K.}
    \label{fig:oracle_attack}
    \vspace{-8pt}
\end{figure}

Fig.~\ref{fig:oracle_attack} shows how our reconstruction improves as K increases in oracle attack~\cite{dusmanu2020privacy}. Still, even with very large K, it is visibly worse than that from direct inversion or the original image. 
For the oracle with NinjaDesc database (last column), the reconstruction is highly privacy-preserving.

\noindent As noted in \cite{dusmanu2020privacy}, an oracle attack is impractical as the attacker does not have access to the original descriptors.
\section{Detailed architectures of the descriptor inversion models}
\label{supp-sec:arch}
\begin{figure*}[ht!]
    \vspace{-0pt}
    \centering
    \includegraphics[width=\linewidth, trim={
        0pt, 0pt, 0pt, 0pt}, clip]{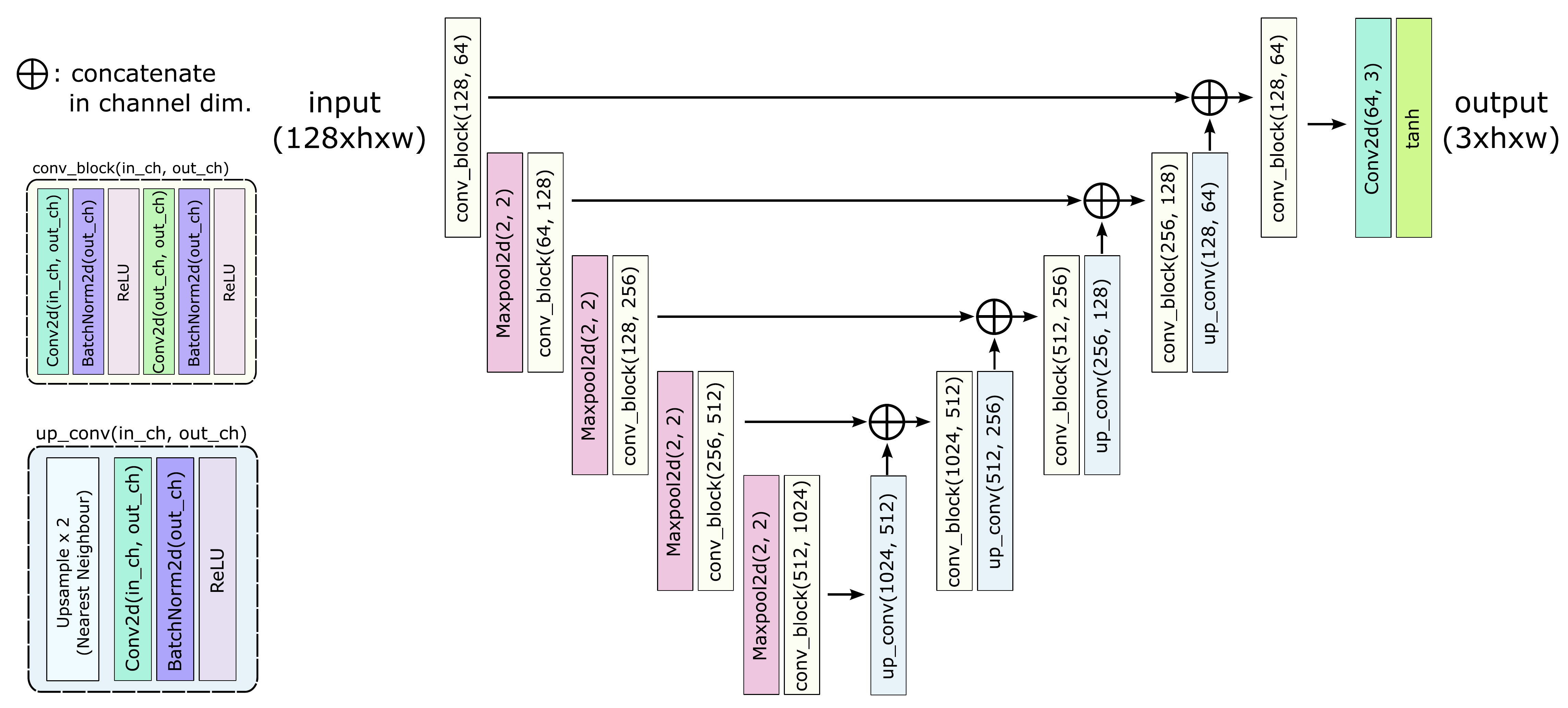}
    \vspace{-0pt}
    \caption{
        UNet Architecture.
    }
    \label{supp-fig:unet_arch}
    \vspace{20pt}
\end{figure*}

\parheader{UNet.} The architecture of the UNet-based descriptor inversion model, which is also used in \cite{dangwal21,invsfm}, is shown in Figure~\ref{supp-fig:unet_arch}.

\begin{figure*}[!ht]
    \vspace{-0pt}
    \centering
    \includegraphics[width=\linewidth, trim={
        0pt, 0pt, 0pt, 0pt}, clip]{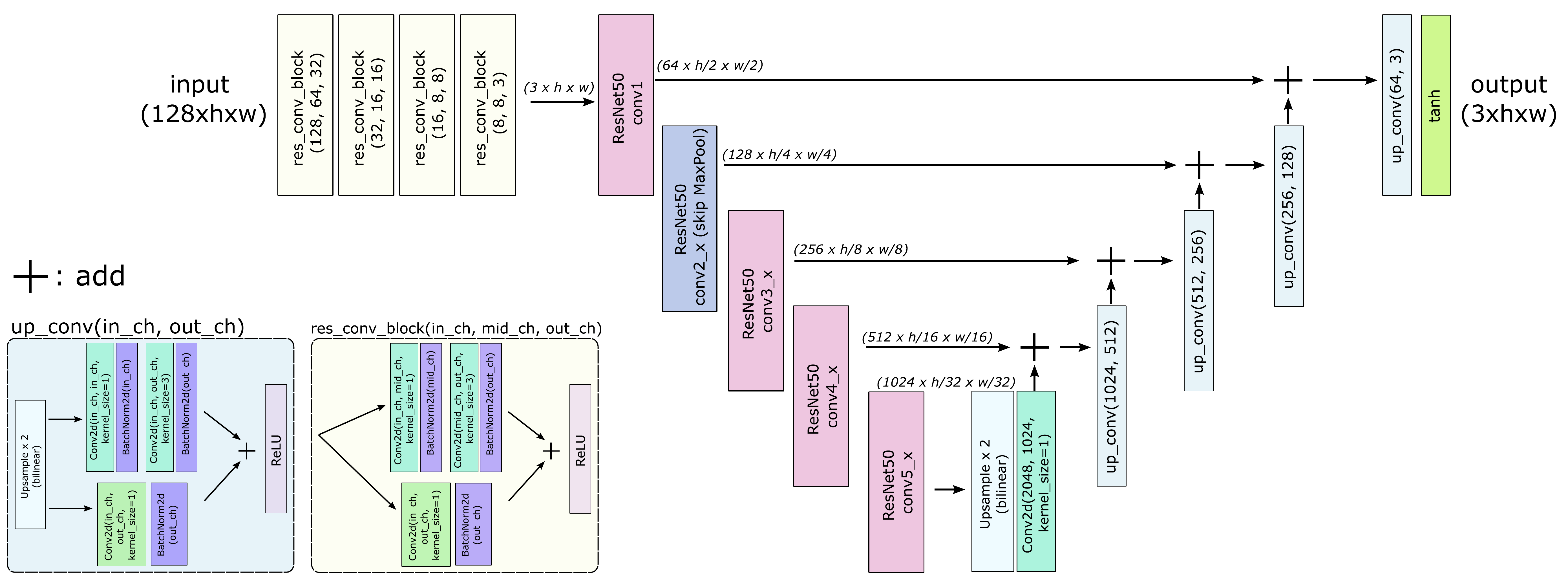}
    \vspace{-0pt}
    \caption{
        UResNet Architecture.
    }
    \label{supp-fig:uresnet_arch}
    \vspace{-0pt}
\end{figure*}

\parheader{UResNet.} Figure~\ref{supp-fig:uresnet_arch} illustrates the architecture of the descriptor inversion model based on UResNet used for the ablation study in the Section 5.2 of the main paper.
The overall ``U'' shape of UResNet is similar to UNet, but each convolution block is drastically different.
We use the 5 stages of ResNet50~\cite{he2016ResNet} (pretrained on ImageNet~\cite{deng2009ImageNet}) \{\texttt{conv1}, \texttt{conv2\_x}, \texttt{conv3\_x}, \texttt{conv4\_x}, \texttt{conv4\_x}\} as the 5 encoding\,/\,down-sampling blocks, except for \texttt{conv2\_x} we remove the \texttt{MaxPool2d} so that each encoding block corresponds to a 1/2 down-sampling in resolution.
Since ResNet50 takes in RGB image as input (which has shape of $3\times h \times w$, whereas the sparse feature maps are of shape $128\times h \times w$), we pre-process the input with 4 additional basic redisual blocks denoted by \texttt{res\_conv\_block} in Figure~\ref{supp-fig:uresnet_arch}.
The up-sampling decoder blocks (denoted by \texttt{up\_conv}) are also residual blocks with an addition input up-sampling layer using bilinear interpolation.
In contrast to UNet, the skip connections in our UResNet are performed by additions, rather than concatenations.
\clearpage

%%%%%%%%% REFERENCES
% {\small
% \bibliographystyle{ieee_fullname}
% \bibliography{references}
% }

% \end{document}

\end{document}